\documentclass[lettersize,journal]{IEEEtran}
\usepackage{amsmath,amsfonts}
\usepackage{algorithmic}
\usepackage{algorithm}
\usepackage{array}
\usepackage[caption=false,font=normalsize,labelfont=sf,textfont=sf]{subfig}
\usepackage{textcomp}
\usepackage{stfloats}
\usepackage{url}
\usepackage{verbatim}
\usepackage{graphicx}
\usepackage{cite}
\usepackage{threeparttable}
\begin{document}

\title{Learning Stage-wise GANs for Whistle Extraction in Time-Frequency Spectrograms}
% Pu Li, Marie A. Roch, Holger Klinck, Saumil Shah, Erica Fleishman, Douglas Gillespie, Eva-Marie Nosal, Yu Shiu, Danielle Cholewiak, Tyler Helble, and Xiaobai Liu

% Saumil Shah, Erica Fleishman, 

\author{Pu Li$^{1, 2}$, Marie A. Roch$^{1}$, Holger Klinck$^{3,4}$, Erica Fleishman$^{5}$, Douglas Gillespie$^{6}$, Eva-Marie Nosal$^{7}$, Yu Shiu$^{8}$, and Xiaobai Liu$^{1}$, ~\IEEEmembership{Member,~IEEE}
\thanks{
% $^{1}$ Computational Science Research Center, San Diego State University
% $^{2}$ Department of Computer Science, University of California, Irvine
% $^{3}$ Department of Computer Science, San Diego State University
% $^{4}$ Center for Conservation Bioacoustics, Cornell University 
% $^{5}$ Marine Mammal Institute, Oregon State University
% $^{6}$ Department of Fish, Wildlife and Conservation Biology, Colorado State University
% $^{7}$ Sea Mammal Research Unit Scottish Oceans Institute, University of St Andrews
% $^{8}$ Department of Ocean and Resources Engineering University of Hawaii 
% $^{9}$ RedRoute, Inc 

$^{1}$ San Diego State University
$^{2}$ University of California, Irvine
$^{3}$ Cornell University 
$^{4}$ Oregon State University
$^{5}$ Colorado State University
$^{6}$ University of St Andrews
$^{7}$ University of Hawaii 
$^{8}$ RedRoute, Inc 

% $^{10}$ Northeast Fisheries Science Center, National Marine Fisheries Service,
% National Oceanic and Atmospheric Administration

% $^{9}$ US Navy Naval Information Warfare Center Pacific
}}

% \author{Marie A. Roch}
% \author{Holger Klinck}
% \author{Saumil Shah}
% \author{Erica Fleishman}
% \author{Douglas Gillespie}
% \author{Eva-Marie Nosal}
% \author{Yu Shiu}
% \author{Danielle Cholewiak}
% \author{Tyler Helble}
% % \author{Xiaobai Liu, ~\IEEEmembership{Member,~IEEE,}}
% \author{Xiaobai Liu, }

% \author{IEEE Publication Technology,~\IEEEmembership{Staff,~IEEE,}
%         % <-this % stops a space
% \thanks{This paper was produced by the IEEE Publication Technology Group. They are in Piscataway, NJ.}% <-this % stops a space
% \thanks{Manuscript received April 19, 2021; revised August 16, 2021.}
% }

% The paper headers
\markboth{Journal of \LaTeX\ Class Files,~Vol.~14, No.~8, August~2021}%
{Shell \MakeLowercase{\textit{et al.}}: A Sample Article Using IEEEtran.cls for IEEE Journals}

% \IEEEpubid{0000--0000/00\$00.00~\copyright~2021 IEEE}
% Remember, if you use this you must call \IEEEpubidadjcol in the second
% column for its text to clear the IEEEpubid mark.

% \settopmatter{printfolios=true}

\maketitle

\begin{abstract}
Whistle contour extraction aims to derive animal whistles from time-frequency spectrograms as polylines. For toothed whales, whistle extraction results can serve as the basis for analyzing animal abundance, species identity, and social activities. During the last few decades, as long-term recording systems have become affordable, automated whistle extraction algorithms were proposed to process large volumes of recording data. Recently, a deep learning-based method demonstrated superior performance in extracting whistles under varying noise conditions. However, training such networks requires a large amount of labor-intensive annotation, which is not available for many species. To overcome this limitation, we present a framework of stage-wise generative adversarial networks (GANs), which compile new whistle data suitable for deep model training via three stages: generation of background noise in the spectrogram, generation of whistle contours, and generation of whistle signals. By separating the generation of different components in the samples, our framework composes visually promising whistle data and labels even when few expert annotated data are available. Regardless of the amount of human-annotated data, the proposed data augmentation framework leads to a consistent improvement in performance of the whistle extraction model, with a maximum increase of 1.69 in the whistle extraction mean F1-score. Our stage-wise GAN also surpasses one single GAN in improving whistle extraction models with augmented data. The data and code will be available at https://github.com/Paul-LiPu/CompositeGAN\_WhistleAugment. 
\end{abstract}

\begin{IEEEkeywords}
Data Augmentation, Generative Adversarial Networks. 
\end{IEEEkeywords}

\section{Introduction}
\subsection{Background}
Spectrograms in the time × frequency domain can show signal structure and are frequently used in audio analysis
\cite{rabiner1993fundamentals}
. Patterns in spectrograms are used for sound event classification \cite{ren2016sound}, bird song recognition \cite{kahl2021birdnet}, music genre classification \cite{lee2009automatic}, automatic music transcription \cite{rizzi2017instrument}, speech emotion recognition \cite{zhang2017speech}, and other tasks. Many acoustic signals have frequency-modulated (FM) components that are visible in spectrograms. Examples include human speech \cite{yost2001fundamentals}, human singing \cite{vijayan2018analysis}, cries of newborns \cite{met2021spectrogram}, vocal melodies \cite{lu2018vocal}, and whale calls \cite{roch2011automated}. In this paper, we concentrate on whistles, the characteristic FM tonal calls of toothed whales.

\begin{figure}[h!]
  \centering
  \includegraphics[width=\linewidth]{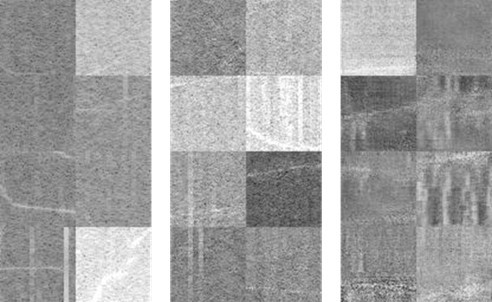}
  \caption{Examples of spectrogram patches of (i) real samples (left); (ii) samples generated by our stage-wise GAN; (iii) samples generated by a single GAN. Multiple 64$\times$64 patches are concatenated for better visualization.}
  \label{fig:whistle_gan_examples}
\end{figure}

Whale calls are used to study species identity \cite{gillespie2013automatic} \cite{jiang2019whistle}, individual identity \cite{janik2013identifying}, behavior \cite{taruski1979whistle} \cite{sjare1986relationship}, communication and social activities \cite{thomsen2002communicative}, and density and abundance \cite{jaramillo2017passive}. Because whistles appear in spectrograms as characteristic contour shapes (Fig.~\ref{fig:whistle_gan_examples} top left), experts can manually recognize animals’ occurrence and label whistles as polylines on spectrograms. Whistle extraction algorithms \cite{roch2011automated} \cite{li2020learning} \cite{mallawaarachchi2008spectrogram} \cite{chen2018sparse} \cite{gruden2020automated} \cite{wang2021method} aim to automate this process and identify each whistle as a polyline. Such extraction is challenging because of the high spatial and temporal variation of ocean sounds. Signals to be analyzed can be affected by recording device characteristics, sea state and propagation conditions, animal behavior, vocalizations from non-target species, and anthropogenic sounds, such as shipping and sonar. 

Traditional methods (e.g., graph search \cite{roch2011automated}) first extract the spectral peaks, i.e., bins with local maximum energy on spectrogram, and then track the trace of whistle signals on the spectrogram by polynomial fitting of peaks \cite{roch2011automated} or probabilistic modeling \cite{gruden2020automated}. Recently, \cite{li2020learning} adapted convolutional neural networks (CNNs) to extract whistles and achieved improved performance. Instead of using spectral peaks, \cite{li2020learning} predicts the confidence associated with the probability that a whistle signal appears in each time × frequency bin, which is similar to semantic segmentation in computer vision. [19] then uses graph search \cite{roch2011automated} to connect bins that are likely to contain a signal. By learning from a large set of annotated samples, the whistle extraction model can recognize noise and whistle patterns, and improve on graph search and probabilistic model results by a large margin. However, the performance of learning-based methods may degrade significantly as the amount of annotated data decreases, and large datasets are not always available because whistle annotation is expensive and time-consuming. This motivates us to explore ways to synthesize whistle data cheaply with existing data by applying learning-based data generation methods.

\subsection{Objectives}
The primary focus of this work was to develop methods that improve whistle extraction models when data are limited, thereby reducing the amount of data annotation required to recognize whistles. Therefore, our experiments mainly addressed situations with few data, and we sought to mitigate the effect of overfitting and improve the model’s transferability for recognizing tonal signals. Although there are many ways to reduce overfitting, e.g., semi-supervised learning \cite{jeong2019consistency} and regularization \cite{li2021learning}, we focused on data augmentation methods for two reasons. First, we seek a method that can be applied to all datasets of frequency-modulated signal, including those  containing no unannotated data. Semi-supervised learning may not be applicable in this scenario. Second, we are interested in characterizing the distribution of whistle data and exploring the effect of novel data on extraction of tonal signals. Regularization terms may not provide insight in this context. We note that our data augmentation method may be combined with a semi-supervised framework or loss function regularization to further improve the system performance. Though it is interesting to have these techniques involved, it is beyond the scope of this work.

Common audio or image data augmentation methods usually transform existing data to acquire new data, e.g., by adding Gaussian noise \cite{he2016deep}, and the augmented samples may implicitly act as a regularizer for the training of deep models \cite{dao2019kernel}. But the distribution of the augmented samples may not be similar to that of the original data; e.g., generated whistle data may have abnormal contour shapes or unrealistic background noise. Previous work \cite{li2020learning} generated novel samples by adding whistle contours to negative samples (background noise that contains no whistle signals), which simulated the situation where the same whistles occur in different ocean environments. However, the generated data did not include novel whistle shapes or background noise patterns, which restricted the variance in the data. 

In this paper, our goal is to generate novel pairs of whistle data and labels. Although changes in noise affect vocalizations of many taxa \cite{brumm2011evolution}, including toothed whales \cite{au1985demonstration}, we make the simplifying assumption that background noise is independent of whistle contours (contour-shape segmentation of whistles, which indicates the location of whistles on spectrogram and the whistles' frequency modulation). On the basis of this assumption, we decouple the synthesis of background noise and whistle contours. The generated whistle contours are used as labels for the model in \cite{li2020learning}. Next, we add generated whistle signals with the desired contour shape to the spectrogram of background noise; i.e., we generate corresponding whistle data for the whistle contour.

We design our whistle generation algorithm as a series of three generative adversarial network (GAN) modules. The first GAN learns the ocean noise environment; it maps random numbers that have a Gaussian distribution to spectrograms representing background noise. The next GAN learns to map random inputs to spectrograms with whistle-like FM sweeps. The third GAN combines the outputs of the first two GAN modules, synthetic background noises and whistles, to obtain a synthetic whistle spectrogram. The generated whistle should follow the whistle contour’s shape in the input. We employ an unpaired domain transfer framework, CycleGAN \cite{zhu2017unpaired}, to learn how synthetic noise and whistles can be merged into a synthetic spectrogram. While the original CycleGAN can generate slightly misaligned whistle signals from the desired contours, we exploit the whistle extraction network learned from annotated data to enforce the bin-wise consistency between generated whistles and input contours. 

Another challenge is that GANs may not learn well with limited data. This may lead to corrupted synthesis, especially of the whistle contours. We observe that corrupted data have less confident predictions: the predicted probability is neither close to 0 nor close to 1, and thus the entropy is high. Accordingly, we introduce a method to prune such low-quality generated samples. Furthermore, because imperfect learning by GANs with few data may lead to discrepancies between the distributions of real data and generated data, we employ auxiliary batch normalization (ABN) layers \cite{xie2020adversarial} which separate the statistics of real and generated data to reduce the possible harmful effect of training with generated data.

\subsection{Contributions}
We made three contributions. First, we proposed the stage-wise composite GANs to generate novel whistle extraction data, including spectrograms and corresponding whistle contour labels. Our experiments showed that the proposed stage-wise GAN surpassed the vanilla GANs with respect to the visual quality of the generated data (Fig.~\ref{fig:whistle_gan_examples} middle and right). Second, we designed a comprehensive strategy to use GAN-generated samples to improve whistle-extraction models. We set criteria to remove corrupted data and we redesigned the whistle extraction network by adding ABN layers to optimize the training with generated data. Third, we applied our proposed data augmentation methods to varied amounts of whistle extraction data and observed consistent and significant improvements. Although GAN frameworks have been used for spectrogram generation and data augmentation in audio recognition tasks \cite{wali2022generative}, to our knowledge, this is the first work to apply GAN-based augmentation to audio spectrogram segmentation data.

% \begin{figure*}[h!]
%   \centering
%   \includegraphics[width=0.8\linewidth]{figs/fig_whistle_gan2.jpg}
%   \caption{Illustration of our proposed framework. In stage 1, we use a generator to map random number vector $z$ to a spectrogram patch of noise. In stage 2, we map random number vector $z$ to a patch of whistle contour. In stage 3, we use another generator to map whistle contour and spectrogram noise to a spectrogram patch containing the input noise and whistle signals.}
%   \label{fig:whistle_gan}
% \end{figure*}

\begin{figure}[ht!]
  \centering
  \includegraphics[width=0.9\linewidth]{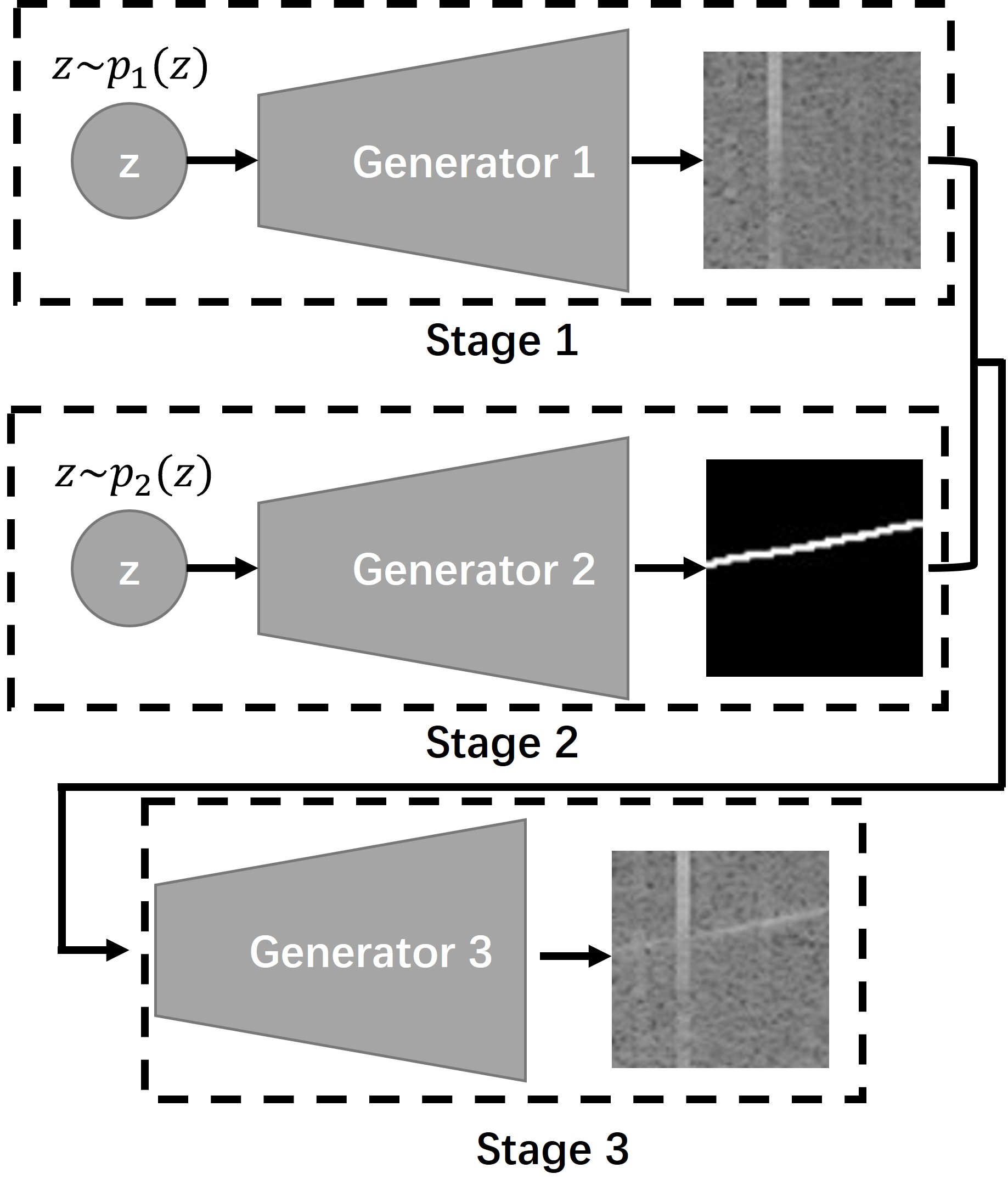}
  \caption{Sketch of the proposed stage-wise GAN frameworks. The first two generators produce a spectrogram patch of background noise and a spectrogram patch of foreground whistle contour, respectively. These patches serve as inputs for the third generator.}
  \label{fig:whistle_gan}
\end{figure}

\section{RELATED WORKS}
\subsection{Whistle Contour Extraction}
There are three main classes of methods for extracting whale frequency-modulated whistles. The first is models that predict the probability of whistle peaks conditioned on past observations. Examples of this class include tests of hypothesized spectrogram region distributions \cite{dadouchi2013automated}, Bayesian inference \cite{halkias2006call}, Kalman filters \cite{mallawaarachchi2008spectrogram}, and Monte-Carlo density filters \cite{roch2011automated} \cite{white2008introduction} \cite{gruden2016automated} \cite{gruden2020automated}. The second class, trajectory-search methods, seeks energy peaks along the frequency dimension and connects those peaks along the time dimension on the basis of trajectory estimation \cite{gillespie2013automatic} \cite{roch2011automated} \cite{mellinger2011method}. Improved trajectory-search methods reduce excessive numbers of false positives by applying ridge regression to local contexts \cite{kershenbaum2013image} or energy minimization algorithms to ridge regression maps \cite{serra2020active}. 

In recent years, the third class, deep learning methods, has been applied to process tonal information. Early works included extraction of information from human speech \cite{han2014neural} and music \cite{bittner2017deep}. Deep neural networks were also applied to toothed whale whistles \cite{jiang2019whistle} \cite{liu2018classification}, but the goal of these works was to classify a time segment to species or call type rather than to extract detailed time × frequency information. In \cite{li2020learning}, we proposed a deep neural network to extract time × frequency contours of individual whistles. We apply our proposed data augmentation system to the training of whistle extraction model developed in \cite{li2020learning}.

\subsection{Generative adversarial networks}
Generative adversarial networks (GANs) are a category of generative models. GANs are widely used for artificial image generation, e.g., face manipulation \cite{ngo2021self}, compression noise removal \cite{galteri2019deep}, and generating images of people \cite{lu2021generate}. We adapted the methods from these computer vision tasks to generate realistic spectrograms that served as our novel training data. The landmark work on GANs \cite{goodfellow2014generative} proposed one generator network that synthesizes samples ($G: X \rightarrow Y$, where $X$ is a random vector and $Y$ is a generated sample) and one discriminator network that learns to distinguish between generated samples and real samples. These networks are coupled in a zero-sum game with each network trying to outperform the other. Following \cite{goodfellow2014generative}, researchers have improved the network architecture of GANs \cite{radford2015unsupervised} and objective functions  \cite{gulrajani2017improved} to stabilize the training of GANs. Those GANs implicitly learn the distribution of real samples, and novel data can be sampled from the distribution. We employ this type of GAN to generate novel spectrogram noises and whistle contours.

Another type of GAN tackles the image-to-image translation problem, aiming to learn a mapping ($F: x \rightarrow y$, where $x\in X$, $y\in Y$) between a source domain $X$ and a target domain $Y$, e.g., transfer a horse in the image to a zebra. CycleGAN \cite{zhu2017unpaired} extends this idea by leaning two mappings ($F: x \rightarrow y$, $G: y \rightarrow x$,  where $x\in X$, $y\in Y$) without the need for pairwise correspondence between the elements of $X$ and $Y$. This idea can be adapted to our task to generate spectrograms containing whistles, where $X$ is the domain containing pairs of desired whistle contours and spectrograms with background noise, and $Y$ consists of spectrograms with whistles and noise. Recent work improved the idea of \cite{zhu2017unpaired} by adding a spatial attention mechanism \cite{emami2020spa} and image quality assessment term \cite{chen2019quality}.

\subsection{GAN-based augmentation}
GANs provide an option to generate novel data by learning the distribution of existing data and sampling data from the distribution, which is a valuable addition to the common augmentation techniques that are based on data transformation. Vanilla GAN models, which map random numbers to generated samples, have been used for data augmentation. 
\cite{frid2018synthetic} trained a GAN model to augment computed tomography (CT) images of livers for the classification of lesions. \cite{mariani2018bagan} applied a conditional GAN to augment samples from given categories and restore the balance of imbalanced image classification data. 
\cite{bowles2018gan} applied progressively grown GANs (PGGANs) to a brain segmentation task, and the generator learned to synthesize the generated sample and corresponding segmentation labels. Domain transfer GANs have also been used for data augmentation. \cite{huang2018auggan} applied CycleGAN to day-to-night image translation, which helped to improve the object detection model.
% \cite{mertes2020data} used the pix2pix \cite{isola2017image} framework to generate carbon fiber defect images from real labels. 

Despite the success of GANs in synthesizing visually appealing samples and augmenting existing data, there are still limitations of GANs for synthesizing high-quality augmented data, especially for pixel-wise regression tasks such as semantic segmentation. First, GANs usually suffer from mode collapses \cite{zeng2021strokegan}: the generated samples may have lower variance than the real samples. Second, GANs may generate samples with artifacts or failure regions \cite{osakabe2021cyclegan}, which may especially hamper the training of pixel-wise regression tasks. A sample selection method may be required to choose high-quality samples from the GAN-generated samples \cite{mun2017generative}. 
Third, the training of GANs can be unstable, which results in different distributions of generated samples and real samples \cite{agarwal2021detecting}. 

Therefore, GAN-based data augmentation usually requires improving the quality of generated samples. A common solution is to use real samples or computer graphics models in the generator network. In \cite{antoniou2017data}, the GAN learned to generate samples conditioned on real samples and random numbers. Similarly, \cite{mu2020learning} transferred synthetic images built by computer graphics models to realistic images, and the augmented samples improved models in estimation of gazes, hand poses, and animal poses. 
Another way to improve training of the GAN is to use supervision from target tasks. \cite{waheed2020covidgan} added an auxiliary classification head on the discriminator of GAN and used the classification loss to guide discriminator and generator learning.

Recently, stage-wise GANs were proposed to augment data for pixel-wise regression tasks. \cite{pandey2020image} employed a two-stage GAN augmentation of cell nuclei segmentation data. Their framework generates a cell nuclei segmentation mask in the first stage and images of nuclei in the second stage. Our proposed method is closely related to \cite{pandey2020image}, and we further separate the learning of object appearance and the segmentation mask. This separation can be extended to other semantic segmentation scenarios. For example, when generating a scene containing road and cars, our framework may first generate the appearance of the road and car independent of the segmentation mask, then generate an image of the scene according to the segmentation mask and the appearance of the objects (road and cars). In this way, our framework explores the distribution of object appearance and provides variance in the appearance of objects in the generated image of the scene. Another improvement is that we employ the knowledge from segmentation networks to regularize bin-wise correspondence between generated samples and labels.

\section{METHODS}
The objective of this work is to develop a data augmentation approach to generate novel data for whistle extraction. We treat the cropped patches from the time-frequency spectrograms as data samples, and we employ stage-wise GANs, which we call WAS-GANs (\textbf{W}histle \textbf{A}ugmentation \textbf{S}tage-wise \textbf{G}enerative \textbf{A}dversarial \textbf{N}etworks), to generate both negative samples (noise only) and positive samples (whistles in the presence of noise). Our techniques can be extended to other acoustic tasks or computer vision tasks, e.g., sound classification and semantic segmentation. 

Fig.~\ref{fig:whistle_gan} illustrates the three stages of our sample generation approach. In Stage 1, a Wasserstein GAN with gradient penalty (WGAN-gp) \cite{gulrajani2017improved} learns to produce the negative samples containing background noises. In Stage 2, we train another WGAN-gp model with the real whistle contour annotations to generate whistle contour segmentation masks. In Stage 3, we use a CycleGAN \cite{zhu2017unpaired} to generate positive samples. The whistle signals are added to the negative samples obtained in Stage 1 according to contour shapes defined in Stage 2. The positive samples and segmentation masks are used as the whistle extraction data and labels, respectively. Both generated negative samples and positive samples are used to train the whistle extraction model, and the resulted whistle extraction performance is used to assess our GAN-based  augmentation.

\subsection{GAN-based negative sample synthesis}
\label{sec:gan_neg}
We assume that the underwater background noise (negative samples) follows an implicit distribution. The generator learns the mapping between a multivariate Gaussian distribution and the distribution of negative samples. While many GAN models can learn this mapping, we chose WGAN-gp because its training is relatively stable \cite{gulrajani2017improved}. The model includes a generator network, $G$, and a discriminator network, $D$. Network $G$ maps a multivariate Gaussian random variable to generate  negative samples. Network $D$ estimates the Wasserstein distance between real samples and generated background noise (negative) samples. We denote $P_r$ as the distribution of real data $x$; $P_g$ as the distribution of generated data implicitly defined by $\widetilde{x}=G(z)$, where $z$ is a random vector following the standard multivariate Gaussian distribution; and $\hat{x}$ as a randomly weighted sum of x and $\widetilde{x}$. The loss function for the discriminator network is defined as
\begin{multline}
\label{eq:eq_loss_neg}
L={\mathbb{E}}_{\tilde{x}\sim {\mathbb{P}}_g\ }\left[D\left(\tilde{x}\right)\right]-{\mathbb{E}}_{x\sim {\mathbb{P}}_r\ }\left[D\left(x\right)\right]+ \\ 
\lambda {\mathbb{E}}_{\hat{x}\sim {\mathbb{P}}_{\hat{x}}\ }\left[{(||{\left.{\mathrm{\nabla }}_{\hat{x}}D\left(\hat{x}\right)\right.}||_2-1})^2\right]
\end{multline}
where $\nabla_{\hat{x}}D\left(\hat{x}\right)$ is the gradient of discriminator $D$'s output on $\hat{x}$. This loss function encourages the discriminator to maximize the estimated Wasserstein distance between real and generated samples. The gradient penalty term ${\mathbb{E}}_{\hat{x}\sim {\mathbb{P}}_{\hat{x}}\ }\left[{(||{\left.{\mathrm{\nabla }}_{\hat{x}}D\left(\hat{x}\right)\right.}||_2-1})^2\right]$ enforces a soft version of Lipschitz constraint on the discriminator network. The loss function for the generator network is
\begin{equation}
\label{eq:eq_loss_neg_g}
L_G={\Bbb{E}}_{z\ }\left[-D\left(G(z)\right)\right]
\end{equation}
which encourages the generator to generate samples that have a small estimated Wasserstein distance from the real samples, i.e., to follow a distribution similar to that of the real data.

\subsection{GAN-based positive sample synthesis}
\begin{figure}[t!]
  \centering
  \includegraphics[width=\linewidth]{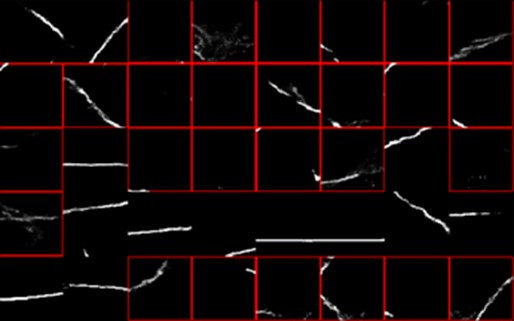}
  \caption {Illustration of whistle contour selection.  Low-quality generated patches are highlighted by red bounding boxes. Multiple 64$\times$64 patches are concatenated.}
  \label{fig:whistle_selection_examples}
\end{figure}

% \begin{figure}[h!]
%   \centering
%   \includegraphics[width=\linewidth]{figs/fig_whistle_tsne.jpg}
%   \caption{Segmentation masks of whistle contours are reduced to 50 dimensions by principal component analysis (PCA) \cite{pearson1901liii} and then reduced to 2 dimensions by t-stochastic neighbor embedding (t-SNE) \cite{van2008visualizing}. The upper left plot shows the distribution of  real whistle contours. The upper right plot shows the distribution of generated whistle contours. When poor-quality exemplars such as those highlighted in Fig.~\ref{fig:whistle_selection_examples} are removed (lower right), the distribution looks more similar to that of the real samples.
% }
%   \label{fig:fig_whistle_tsne}
% \end{figure}

We split synthesis of positive samples (spectrograms containing whistles) into two stages: generation of whistle contours and injection of the whistle into synthetic background noise. In the first stage, we employ the same networks and loss functions as in Section~\ref{sec:gan_neg}, given the assumption that the shape of whistle contours is independent of the underwater environments. 

In the second stage, we aim to generate positive samples according to the synthetic background noise and whistle contours. We treat this as an unpaired domain transfer task, which can be solved effectively by CycleGAN \cite{zhu2017unpaired}. Our source domain, A, consists of pairs of negative samples and whistle contours, and the target domain B includes positive samples. We adopt the CycleGAN from \cite{zhu2017unpaired} for our experiments, but any improved model readily can be used in our framework. 

There are two sets of generator and discriminator networks in CycleGAN. $G_A$ denotes the generator network that transfers samples from domain $A$ to domain $B$, i.e., generates whistle with the desired shape on the background noise spectrogram. $D_A$ denotes the discriminator network that distinguishes between real and generated spectrograms in domain $B$. $G_B$ is the network that transfers samples from domain $B$ to domain $A$, effectively separating the whistle contour from the background noise. Because we assume that the whistle contour and background noise are independent, we do not use a single $D_B$ network for the joint distribution of whistle contours and background noise. Instead, we use two $D_B$ networks for the marginal distributions, one to discriminate negative samples and one to discriminate whistle contours. 

Instead of directly generating positive samples by $G_A$, we let $G_A$ predict a residual term (whistle signals without background noises) to be added to the negative samples. By denoting a negative sample as $I_N$, a whistle contour as $I_W$, and the generated positive sample as $I_P’$, the process can be described as 
\begin{equation}\label{eq:eq_pos}
I_P'=I_N+\gamma G_A(I_N,\ I_W)
\end{equation}
where $\gamma$ is a factor that controls the signal strength and accounts for variability in the received signal level. This parameter can simulate the variation in signal strength caused by variation in signal source strength or the distance between the animal and recording devices.

To enforce the bin-wise correspondence between generated positive samples and whistle contours, i.e., to avoid misalignment between generated whistle extraction data and labels, we use the whistle extraction models, which are trained on the same set of real samples as CycleGAN, to design a regularization term for $G_A$  training. We call this term a loss function for the pixel-wise consistency, and represent it as
\begin{equation}\label{eq:eq_loss_consistency}
L_{consistence}=||f(I_P')-I_W||_1
\end{equation}
where $f$ denotes the whistle extraction model and $f(x)$ is the model’s output, a confidence map indicating the presence of whistle energy in each bin of the spectrogram, with an input $x$. This loss encourages the whistle signals to appear at the same position as the desired whistle contour.

To guarantee that the generated positive samples have the same background magnitude as the input negative samples, we also include the identity loss,
\begin{equation}\label{eq:eq_loss_identity}
L_{identity}=||G_A(I_N, 0)||_1+||G_B(I_N)-(I_N,0)||_1
\end{equation}
where $0$ indicates an empty whistle contour input, i.e., we do not want the CycleGAN to generate any whistles. We denote $(I_N,0)$ as the concatenated $I_N$ and empty whistle segmentation map. $G_A$ should produce residuals of zero when there are no input whistle contours. We also use adversarial loss, $L_{D_A}$, $L_{D_B}$, $L_{G_A}$, $L_{G_B}$, and cycle consistence loss ($L_{cyc}$) from CycleGAN
\begin{equation}\label{eq:eq_loss_cycle_da}
% L_{D_A}={({{D}}_{{A}}({{I}}_{{P}})-1)}^2+{({{D}}_{{A}}({{G}}_{{A}}({{I}}_{{N}},\ {{I}}_{{W}})))}^2
L_{D_A}={({{D}}_{{A}}({{I}}_{{P}})-1)}^2+{({{D}}_{{A}}(I_P'))}^2
\end{equation}
\begin{equation}\label{eq:eq_loss_cycle_db}
L_{D_B}={({{D}}_{{B}}({{I}}_{{N}},\ {{I}}_{{W}})-1)}^2+{({{D}}_{{B}}({{G}}_{{B}}({{I}}_{{P}})))}^2
\end{equation}
\begin{equation}\label{eq:eq_loss_cycle_ga}
% L_{G_A}={({{D}}_{{A}}({{G}}_{{A}}({{I}}_{{N}},\ {{I}}_{{W}}))\ -\ 1)}^2
L_{G_A}={({{D}}_{{A}}(I_P')\ -\ 1)}^2
\end{equation}
\begin{equation}\label{eq:eq_loss_cycle_gb}
L_{G_B}={({{D}}_{{B}}({{G}}_{{B}}({{I}}_{{P}}))-\ 1)}^2
% L_{G_B}={({{D}}_{{B}}(I_P')-\ 1)}^2
\end{equation}
\begin{multline}
\label{eq:eq_loss_cycle_consistency}
% L_{cyc}={||G_B(G_A\left({{I}}_{{N}},\ {{I}}_{{W}}\right))-\left({{I}}_{{N}},\ {{I}}_{{W}}\right)||}_1+ \\
% {||G_A(G_B\left({{I}}_{{P}}\right))-{{I}}_{{P}}||}_1
L_{cyc}={||G_B(I_P')-\left({{I}}_{{N}},\ {{I}}_{{W}}\right)||}_1+ \\
{||G_A(G_B\left({{I}}_{{P}}\right))-{{I}}_{{P}}||}_1
\end{multline}
where $I_P$ refers to real positive samples. We simplify the notation of two $D_B$ networks in one $D_B$ function in the above equation. The full objective for generators is 
\begin{equation}\label{eq:eq_loss_cycle_g}
L_G=L_{G_A}+{L_{G_B}\ + \lambda }_0L_{cyc}+{\lambda }_1L_{consistence}+{\lambda }_2L_{identity}
\end{equation}
where $\lambda_0$, $\lambda_1$, and $\lambda_2$ control the relative importance of the corresponding loss items. The full objective of the discriminator is
\begin{equation}\label{eq:eq_loss_cycle_d}
L_D=L_{D_A}+L_{D_B}
\end{equation}
Ideally, $D_A$, $D_B$ will assign 1 to real samples and assign 0 to generated samples with this training objective. $G_A$, $G_B$ will try to fool the discriminators and generate realistic samples.

\subsection{Whistle extraction model}
We use the whistle extraction model from \cite{li2020learning} as our baseline. This model, which is similar to a selective edge detection model, produces a confidence map of the whistle signals. Although the generated samples are visually similar to real samples (Fig.~\ref{fig:whistle_selection_examples}), 
the distributions of the real and generated whistle contour may differ due to the imperfect training of GAN when data are limited. 
% (Fig.~\ref{fig:fig_whistle_tsne}). 
This discrepancy decreases the accuracy of our whistle extraction model when we use the generated samples for data augmentation. Therefore, we use ABN layers \cite{xie2020adversarial}; i.e., we use auxiliary BatchNorm (BN) layers for forwarding generated samples and normal BN layers for real samples. We share the same convolutional layers for real and generated samples. By denoting the input sample as $x$, the whistle signal label as $y$, and the whistle extraction model as $f$, the loss without ABN can be described as 
\begin{equation}\label{eq:eq_loss_whistle}
L=||y-f(x))||_2
\end{equation}
The loss with ABN is
\begin{multline}
\label{eq:eq_loss_abn}
L=\frac{1}{(1+\lambda)}(||(y_{real}-f(x_{real}))||_2+ \\
\lambda {||y_{fake}-f_{abn}(x_{fake}))||}_2)
\end{multline}

where $x_{real}$, $y_{real}$ are the real samples and labels, respectively, and $x_{fake}$, $y_{fake}$ are the generated samples and labels, respectively. $\lambda$ is a factor to adjust the weights of real data and generated data in loss calculation. $f_{abn}(x)$ denotes the output of the whistle extraction model for input $x$ when the auxiliary BN layer is used in forwarding. We empirically find that ABN layers improve the whistle extraction performance when the distributions of the generated and real samples may be different.

The quality of GAN-synthesized samples is affected by the number of real samples available for training. The generator may synthesize poor-quality samples when the number of real examples used in GAN training is low. Fig.~\ref{fig:whistle_selection_examples} provides examples of synthetic whistle contours when 2500 real positive samples are used for GAN training, including whistle contours that are of poor quality. Therefore, we designed two heuristic conditions for selecting high-quality generated samples. Denoting the value of an individual bin in the whistle contour patch as p, we select the sample for training the whistle extraction model when
\begin{equation}\label{eq:eq_thres_entropy}
\sum{-p{log p\ }}<T_e
\end{equation}
and
\begin{equation}\label{eq:eq_thres_conf}
\sum{\delta (p-T_c)}>T_p
\end{equation}
where
\begin{equation}\label{eq:eq_select}
\delta(x)=\left\{\begin{matrix}
0 & x \le 0\\ 
1 & x >  0
\end{matrix}\right.
\end{equation}
$T_e$ is a threshold for the sum of the pixel entropy, so the first condition removes generated whistles with diffuse medium-intensity signals (high entropy). The second condition chooses samples in which more than $T_p$ bins have intensity above $T_c$, allowing samples with short whistle fragments to be removed.

\begin{figure}[t!]
  \centering
  \includegraphics[width=\linewidth]{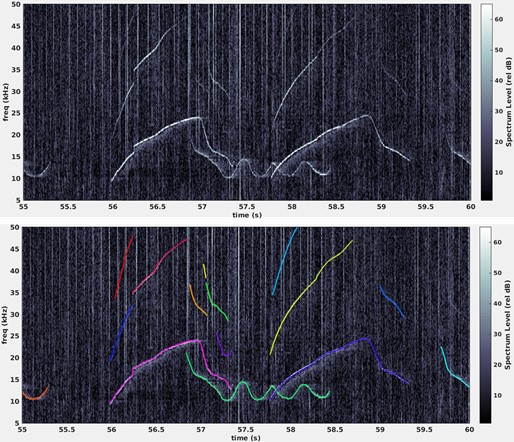}
  \caption {Illustration of whistle extraction. (Top) spectrogram visualized by \textit{Silbido} \cite{roch2011automated}; (Bottom) extracted whistles, where each whistle is highlighted with a different color.}
  \label{fig:fig_whistle} 
\end{figure}

\section{DATA AND IMPLEMENTATION}
\subsection{Datasets}
We used the whistle extraction data from the 2011 workshop on detection, classification, localization, and density estimation of marine mammals (DCLDE 2011, available on the MobySound Archive \cite{mellinger2006mobysound}). These data contain recordings of calls made by five toothed whale species: long-beaked common dolphins (\textit{Delphinus capensis}), short-beaked common dolphins (\textit{Delphinus delphis}), bottlenose dolphins (\textit{Tursiops truncatus}), melon-headed whales (\textit{Peponocephala electra}), and spinner dolphins (\textit{Stenella longirostris}). Whistle contours were annotated by trained analysts across the 5-50 kHz bandwidth as described in \cite{roch2011automated}. We use 30 recordings from the 5 species to train and 12 recordings from 4 species to test. Short-beaked common dolphins are removed from evaluation because some of the files had annotation errors. The training data consisted of approximately 127 min of recordings with 12,539 annotated whistles. The test data ($\sim$43 min of acoustic data) contained 6,011 annotated whistles.

\begin{figure}[t!]
  \centering
  \includegraphics[width=0.7\linewidth]{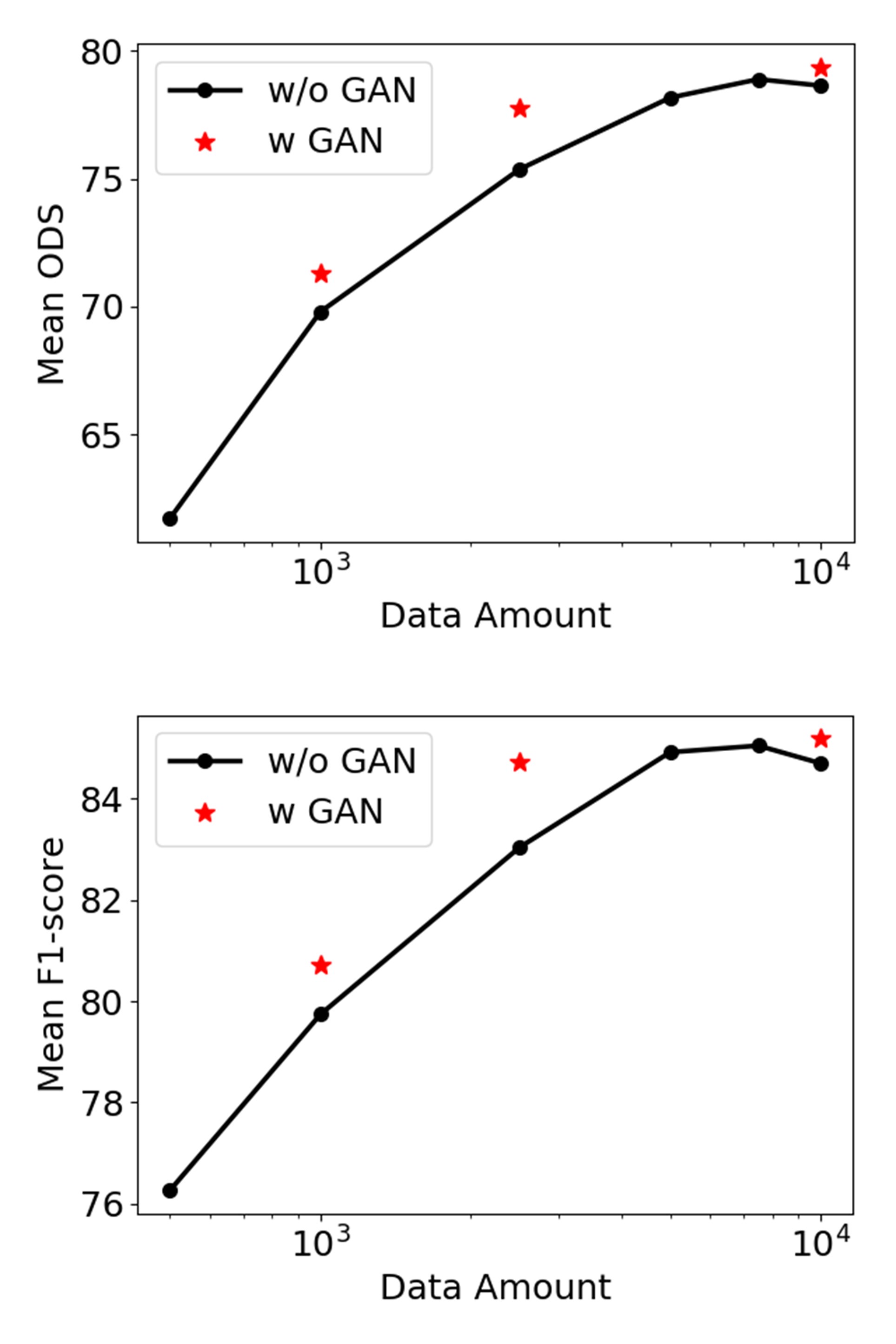}
  \caption{Mean spectral peak detection F1-score (upper) or mean whistle extraction F1-score (lower) against the number of real positive samples in the training set. Optimal Dataset Scale (ODS) is an edge detection metric that assesses peak detection. "w/o GAN" and "w GAN" indicates the performance without and with GAN augmentation, respectively. }
  \label{fig:whistle_perform_curve}
\end{figure}

We computed log-magnitude spectrograms for the whistle extraction model and the GAN-based data synthesis. We employed series of discrete Fourier transforms in spectrogram computation. 8 ms Hamming-windowed frames (125 Hz bandwidth) were computed every 2 ms, and we empirically restricted the dynamic range of the $log_{10}$ magnitude spectrogram to the range [0, 6] (an intensity range of 0 to 120 dB rel.), i.e., we transformed the values $<$0 to 0, and those $>$6 to 6. We divided the spectrogram values by 6 which made them within [0, 1], and discarded the spectrogram values outside of the annotation frequency range of 5-50 kHz (361 frequency bins), which covers the frequency range of most delphinid whistles and their harmonics. 

For network training, we partitioned the spectrogram into 64 $\times$ 64 patches, where each patch covered a time interval of 128 ms and frequency interval of 8 kHz. For the training data, we selected the positive patches with a sliding window with a 25 pixel step size across portions of spectrograms containing whistles, which led to 115,968 positive patches available for training. We randomly selected the same number of negative patches, which only contain noise, and combined them with positive patches as our training data (referred to as the full dataset). Most of our experiments used a subset of the full data (referred to as a reduced dataset). We describe the details of generating the reduced dataset in Section~\ref{sec:varied_data}. 

\subsection{Networks and Algorithms}
\subsubsection{Whistle extraction network}
We used the same network architecture as \cite{li2020learning}. The model has 10 convolutional layers, including 1 input layer, 4 residual blocks (each block contains two convolution layers), and 1 output layer. The input layer and output layer use kernel size 5 and padding size of 2, and other layers use a kernel size of 3 and a padding size of 1. All hidden layers have 32 channels. The model input is a one-channel spectrogram and the output is a confidence map of whistle occurrence. The size of the output confidence map is the same as that of the input spectrogram.

We trained the whistle extraction model with an Adam optimizer (initial learning rate=$1\times{10}^{-3}$, betas = [0.9, 0.999], weight decay=$5\times{10}^{-4}$) for $1\times{10}^6$ and $3\times{10}^5$ iterations on the full dataset and reduced datasets, respectively. The learning rate was multiplied by 0.1 every $4\times{10}^5$ and $1\times{10}^5$ iterations for the full and reduced datasets, respectively. We set the batch size to 64, and we used 64 real samples and 64 generated samples in each iteration for data augmentation experiments. We used $\lambda$=1 in the loss function of Eq.~\ref{eq:eq_loss_abn} for our experiments with generated data, which make the generated samples have the same contribution of loss as real examples.

\subsubsection{WGAN}
\label{sec:wgan}
We used the same WGAN architecture for the generation of whistle contours and negative samples. 
The generator network uses a fully-connected layer to output feature maps of size (512,4,4) from a 128-dimensional standard Gaussian distribution. Four groups of convolutional layers and pixel shuffle layers are used to gradually enlarge the feature map to $64\times64$. A Tanh layer is used to output the $64\times64$ patch. The discriminator network takes the generated samples and real samples as input, and outputs the Wasserstein distance estimation. It contains 4 convolutional layers with a stride of 2 and a fully connected layer. The networks are optimized by Adam optimizers (initial learning rate = $1\times{10}^{-4}$, betas = [0.5, 0.9], batch size = 64) for $3\times{10}^4$ and $5\times{10}^4$ iterations on the reduced and full datasets, respectively. In each WGAN training iteration, the discriminator is optimized for 5 steps while the generator is optimized for 1 step, where the network parameters are updated by applying the optimizer to one mini-batch of data in each step. For sample selection, we used $T_e$=70, $T_c$=0.5, $T_p$=64.

\begin{figure*}[t!]
  \centering
  \includegraphics[width=0.9\linewidth]{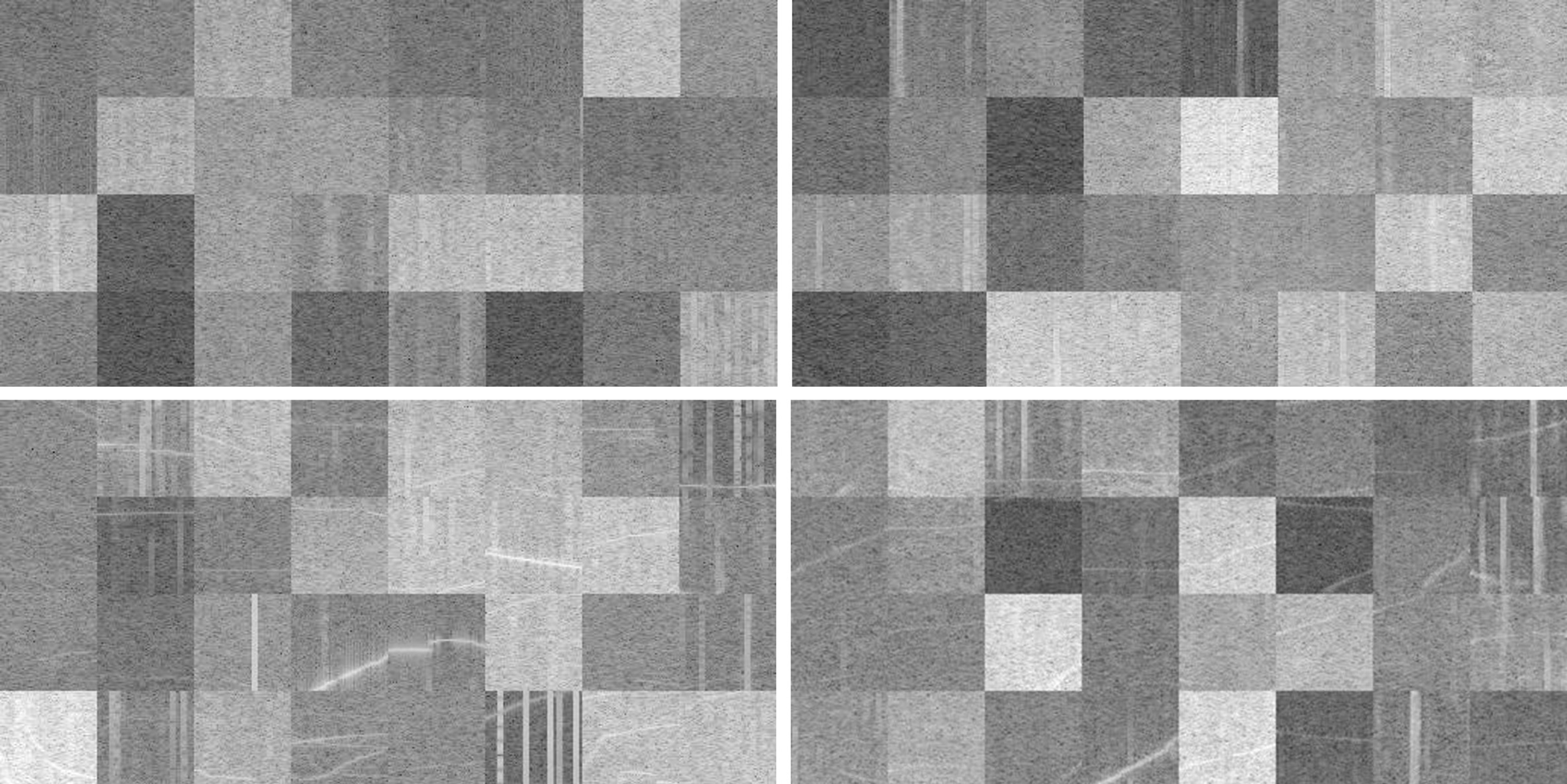}
  \caption{Real background noise samples (upper left); Our GAN generated background noise samples (upper right); Real whistle samples (bottom left); Our GAN generated whistle samples (bottom right). Multiple 64 $\times$ 64 patches are concatenated in each category for better visualization of the data variance.}
  \label{fig:whistle_gan_examples_detail}
  \vspace{-3mm}
\end{figure*}

\subsubsection{CycleGAN}
The GAN model that we used to add whistles on synthetic noise employs the CycleGAN architecture of \cite{zhu2017unpaired}. The generators follow the U-Net \cite{ronneberger2015u} architecture, which has 6 U-Net blocks with a basic width of 64. InstanceNorm layers are used in the U-Net blocks. The discriminator is a fully convolutional network with 3 convolutional layers. We trained the generators and discriminators with Adam optimizers (learning rate = $2\times10^{-4}$, betas = [0.5, 0.999], batch size = 64) for 25,120 iterations (160 epochs for 10,000 real positive samples) for the reduced dataset and 50 epochs for the full dataset. We set $\lambda_0$=10, $\lambda_1$=0.5, and $\lambda_2$=0.5 for Eq.~\ref{eq:eq_loss_cycle_g}. We apply a random $\gamma$ following a unified distribution between (0.5, 1.5) in Eq.~\ref{eq:eq_pos}.

\subsubsection{Graph Search}
We adapted the graph search \cite{roch2011automated} algorithm to the outputs of the whistle extraction network to predict individual whistles. This algorithm maintains sets of graphs, the nodes of which indicate the trace of predicted whistle contours. Multiple crossing whistles can be represented by a single graph.  At each time step, local maximum points (peaks) on the confidence map are selected along the frequency dimension, and peaks with confidence greater than 0.5 are retained as candidate points. For each candidate point, the algorithm either initiates a new graph or extends terminating nodes of existing graphs. Extensions are made when the new node is along a reasonable trajectory predicted by a low-order polynomial fit of the graph path near a terminating node. Graphs that have not been extended within a specified time are considered closed. Closed graphs are removed from the current graph set. When a graph is of a shorter duration than a settable minimum whistle duration, it is discarded. Otherwise individual whistles are extracted from the graph on the basis of an analysis of graph vertices.
% (trajectory crossings)

\vspace{-1mm}

\subsection{Metrics}
\label{sec:metrics}
\subsubsection{Evaluation of confidence maps}
%We first assessed the quality of the whistle-energy confidence maps predicted by the whistle-extraction model. We utilized the BSDS 500 \cite{arbelaez2010contour} benchmark tools and protocol to calculate the highest dataset-scale F1-score over different thresholds (Optimal Dataset Scale, or ODS). We thinned the ground-truth confidence maps to 1 pixel wide and compared them with a predicted confidence map binarized by
% Pu: change 30 to 50. 
%50 
%thresholds evenly distributed between (0, 1). All default parameters in the benchmark tool are used in our evaluation. 

We first assessed the quality of the whistle-energy confidence maps predicted by the whistle-extraction model. To do this, we utilized the BSDS 500 benchmark tools and protocol \cite{arbelaez2010contour} to calculate the highest dataset-scale F1-score across various thresholds (referred to as the ``Optimal Dataset Scale," or ODS). We thinned each ground-truth whistle to a width of one pixel and compared them to predicted confidence maps that were binarized using 50 evenly distributed thresholds between 0 and 1. All default parameters within the benchmark tool were used in our evaluation.

\subsubsection{Evaluation of whistle extraction}
We used  \textit{Silbido} \cite{roch2011automated} to evaluate the quality of whistle extraction after the graph search was applied to the confidence map. This library calculates recall, the percentage of validated whistle contours that were detected; and precision, the percentage of detections that were correct. Then we calculated the precision, recall, and F1-score on testing files of each species and averaged them among all species. We determined the success or failure of whistle extraction results by examining the set of expected analyst annotations as described in \cite{roch2011automated}. We checked whether any of the detections overlapped with the analyst-annotated whistle contour in time. If so, we examined whether each overlapping detection matched the analyst’s annotation. When the average deviation in frequency between the detected contour and annotation was $<$ 350 Hz and the analyst detections had lengths $\geq$ 150 ms, with a signal-to-noise ratio $\geq$ 10 dB over at least 30\% of the whistle, we classified the overlapping detections as matched detections. When an annotated whistle did not meet the above criteria (too short or low intensity), we discarded any matching detections, and they did not contribute to the metrics. We classified unmatched detections as false positives.

\begin{table*}[t!]
\centering
\begin{threeparttable}
\caption{Performance of whistle extraction }
\label{tab:gan}
\begin{tabular}{c|cc|cc|cc|cc}
\hline
N\tnote{a}   & \multicolumn{2}{c|}{Mean ODS}                & \multicolumn{2}{c|}{Mean F1-score}           & \multicolumn{2}{c|}{Mean Precision}          & \multicolumn{2}{c}{Mean Recall}              \\ \hline
      & \multicolumn{1}{c|}{w/o GAN\tnote{b}}    & w GAN      & \multicolumn{1}{c|}{w/o GAN}    & w GAN      & \multicolumn{1}{c|}{w/o GAN}    & w GAN      & \multicolumn{1}{c|}{w/o GAN}    & w GAN      \\ \hline
1000  & \multicolumn{1}{c|}{69.80±2.41\tnote{c}} & 71.33±2.58 & \multicolumn{1}{c|}{79.74±2.94} & 80.73±1.89 & \multicolumn{1}{c|}{85.01±3.54} & 76.86±3.99 & \multicolumn{1}{c|}{84.82±3.44} & 78.32±3.58 \\ \hline
2500  & \multicolumn{1}{c|}{75.37±1.50} & 77.78±0.89 & \multicolumn{1}{c|}{83.04±1.10} & 84.73±0.90 & \multicolumn{1}{c|}{85.88±1.92} & 86.38±1.77 & \multicolumn{1}{c|}{81.29±2.26} & 83.72±1.32 \\ \hline
10000 & \multicolumn{1}{c|}{78.64±0.67} & 79.38±0.38 & \multicolumn{1}{c|}{84.70±1.11} & 85.21±0.80 & \multicolumn{1}{c|}{87.55±1.52} & 87.13±1.86 & \multicolumn{1}{c|}{82.67±1.52} & 83.85±1.13 \\ \hline
all   & \multicolumn{1}{c|}{80.85±0.23} & 81.23±0.10 & \multicolumn{1}{c|}{87.42±0.44} & 87.88±0.14 & \multicolumn{1}{c|}{89.27±0.20} & 89.60±0.31 & \multicolumn{1}{c|}{86.04±0.67} & 86.63±0.36 \\ \hline
\end{tabular}
\begin{tablenotes}
    \item[a] We denote the number of real positive samples for whistle extraction model and GAN training as N; “all” indicate that the full dataset is used. 
    \item[b] w GAN and w/o GAN  indicate the performance of whistle extraction model with or without our GAN generated samples, respectively. The whistle extraction model is the same as \cite{li2020learning}. 
    \item[c] We conduct repeated experiments for each setting, and we report performance average ± standard deviation for each metric. Refer to Section~\ref{sec:metrics} for more details.  
\end{tablenotes}
\end{threeparttable}

\end{table*}

\begin{figure*}[t]
  \centering
  \includegraphics[width=0.9\linewidth]{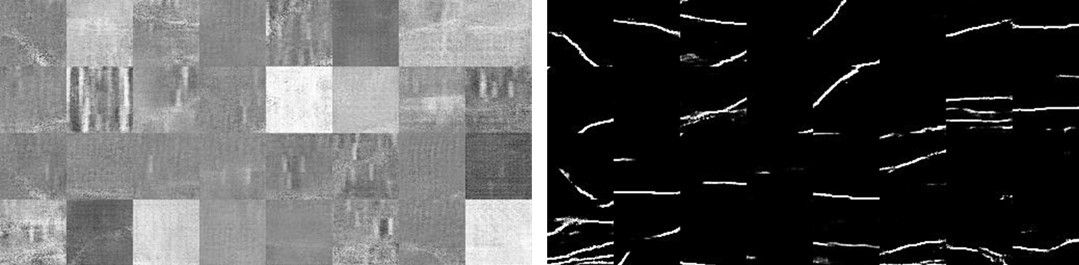}
  \caption{Positive samples (left) and corresponding whistle contour (right) generated by vanilla GAN. Multiple 64 $\times$ 64 patches are concatenated. }
  \label{fig:fig_whistle_vanilla}
\end{figure*}

\section{EXPERIMENTS AND RESULTS}
\subsection{Varied number of annotated samples}
\label{sec:varied_data}
We first studied the effect of varying the amount of training data for our whistle extraction network. Because annotation is expensive, a key motivation for data augmentation is to reduce the number of annotations required. Training effective deep-learning models requires a considerable amount of high-quality annotated data \cite{althnian2021impact}. For the whistle extraction task in this paper, it remains unclear how the whistle models perform when the amount of annotated data varies. To address this issue, we conducted 6 experiments that selected n positive patches and n negative patches, where n = 500, 1000, 2500, 5000, 7500, or 10000. Random selection of patches was structured to ensure that smaller datasets were subsets of larger ones. We repeated this process five times to obtain 5 datasets for each n. For each dataset, we trained whistle extraction models 5 times, and report average performance.

The experimental results are shown in Fig.~\ref{fig:whistle_perform_curve}. The black curves show the performance of the confidence map (ODS) and whistle extraction (F1-score) (upper and lower plots, respectively) with respect to the quantity of training data. While the ODS quantifies the performance of the whistle extraction model in detecting the presence and shape of the whistles, the results suggest that, with more training data, the average ODS increases. The increase in whistle extraction F1-score follows the same trend as ODS. Our results show that increasing the amount of annotated data substantially improves the performance of whistle detection. At the same time, as the amount of data increases, the rate of performance improvement decreases, which means that exponentially more data may be needed to increase performance by 1 unit when the initial dataset is larger. 

\subsection{Data augmentation}
\label{sec:data_aug_results}
We also studied the effect of varying dataset size on GAN training and data augmentation. In this set of experiments, we applied the proposed augmentation method to augment n = 1000, 2500, and 10000 positive samples and negative samples. In each experiment, we generated 10 $\times$ n samples with our WAS-GAN. All GAN networks were randomly initialized and trained once per dataset. For each augmented dataset, we trained the whistle extraction model with ABN for 5 times. 

Fig.~\ref{fig:whistle_gan_examples_detail} shows examples of samples generated by our WAS-GAN (n = 2500). By visually comparing the real samples and generated samples, we see that the noise patterns and whistle signal patterns are well simulated by our GAN networks, e.g., the clicks (wide vertical band of high energy across the frequency domain) are simulated well, as are the width and strength of whistle signals. 

Table~\ref{tab:gan} reports the experiment's ability
to correctly predict time-frequency peaks associated with whistles (mean ODS) and to correctly extract whistles from these predictions (mean F1-score). 
Consistent performance improvements were obtained for both measures. Our methods obtained gains of 1.53, 2.41, and 0.74 in mean ODS, and 0.99, 1.69, and 0.51 in mean F1-score for the three augmentation experiments when n=1000, 2500, and 10000 training patches, respectively. We also obtained improvements of 0.38 and 0.46 in the mean ODS and mean F1-score, respectively, when we used WAS-GAN on the full dataset. 
% As we discuss in Section~\ref{sec:varied_data}, if our augmentation method is not applied, tens of thousands more annotated samples may be needed for similar improvement. 
% Pu: Add more explanation to the calculation. 
% Compared to experiments of n=10000, we use over 100 thousand more annotated samples without augmentation while we increase the whistle extraction F1-score by 2.72. 
% Without our GAN-generated samples, if we wish to increase the F1-score by 0.46 with more human annotation on top of our current dataset, we may need to annotate tens of thousands more samples. 
%
In comparison to experiments using n=10000, we utilized over 100,000 additional annotated samples in our full dataset experiment. These samples were manually labeled as opposed to our GAN augmented samples, and this led to an increase of 2.72 in the whistle extraction F1-score. Without our GAN-generated samples, in order to achieve a 0.46 increase in the F1-score by adding more human-annotated samples to our current dataset, we would have to annotate tens of thousands more samples.
The training stability was notably improved (with a reduction in the variance of the F1 metrics) with the addition of the generated data. These improvements highlight the effectiveness of our proposed stage-wise, GAN-based data augmentation method: the use of augmented data improves spectral peak detection results, which in turn also improves whistle contour extraction results.

%Compared to experiments with n=10000, we use over 100,000 more annotated samples in the full dataset experiment. These samples were manually labeled as opposed to our GAN augmented samples, and increased the whistle extraction F1-score by 2.72. Without our GAN-generated samples, if we wish to increase the F1-score by 0.46 by adding more human-annotated samples to our current dataset, we may need to annotate tens of thousands more samples.
%Training stability was mostly improved (lower variance of the F1 metrics) with the addition of generated data. These improvements show the effectiveness of our proposed stage-wise, GAN-based data augmentation method: use of augmented data improves spectral peak detection results, which also improves whistle contour extraction results. 

\subsection{Ablation study}
We conducted a set of ablation experiments to examine the contributions of different components of the proposed method. We chose datasets with n = 2500 samples for these experiments. The quantitative results are shown in Table~\ref{tab:gan_ablation}. 

\begin{table}[b!]
\centering
\begin{threeparttable}
\caption{ablation study}
\label{tab:gan_ablation}
\begin{tabular}{ccc}
\hline
Experiments         & Mean ODS & Mean F1-score \\ \hline
2500+GAN\tnote{a}            & 77.78    & 84.73         \\
- residual\tnote{b}               & -1.43    & -1.44         \\
- select            & -1.21    & -1.44         \\
- ABN                & -0.68    & -0.98         \\
- ABN, - select      & -0.86    & -1.97         \\
- residual, - select, - ABN & -2.01    & -5.21         \\
vanilla   GAN\tnote{c}       & -0.57    & -1.04   \\
Random Addition\tnote{d}                 & -0.36    & -0.65                   \\
Random Addtion + Gaussian Blur\tnote{e}  & -0.37    & -0.67 \\
\hline

\end{tabular}
\begin{tablenotes}
    \item[a] GAN augmentation from 2500 real positive samples and 2500 negative samples. We report the  whistle extraction performance with our proposed GAN method in this row and the change of performance compared to this row in the following rows. 
    \item[b] -XXX means that component XXX is removed. The components include: (i) residual: residual learning; (ii) select: selection of synthetic whistles with entropy and duration criteria; (iii) ABN: auxiliary batch normalization. 
    \item[c] We replace stage-wise GANs with a single WGAN-gp \cite{gulrajani2017improved} for sample synthesis. 
    \item[d] We remove the third GAN model (CycleGAN) and directly add the output of the first two GANs with random weights. 
    \item[e] We apply random Gaussian blurring to the generated whistle contour before it is added to background noise. 
\end{tablenotes}
\end{threeparttable}
\end{table}

\subsubsection{Residual learning}
In this ablation experiment, we trained the CycleGAN in stage 3 to directly generate positive samples rather than adding the residual to the negative samples (Eq.~\ref{eq:eq_pos}). While we can change the whistle signal magnitude by altering the weight in Eq.~\ref{eq:eq_pos} when the generator outputs residual, the whistle signal’s magnitude is determined by the generator model in this setting. In contrast to the proposed WAS-GAN, we observed a decrease of 1.43 in mean ODS and a decrease of 1.44 in mean whistle extraction F1-score when we removed residual learning. This performance drop might be caused by the fact that the GAN needed to output background noise, which might increase the difficulty and instability of learning. Moreover, the variance of generated data decreases when the magnitude of whistle signals cannot be adjusted by the multiplier of the residual. 

\subsubsection{Patch selection}
This ablation experiment removed the quality assurance filter (Eq.~\ref{eq:eq_thres_entropy} and Eq.~\ref{eq:eq_thres_conf}) for whistles generated by the GAN. As a result, generated whistles similar to those surrounded by the red bounding boxes (Fig.~\ref{fig:whistle_selection_examples}) were included in the training data. The mean ODS dropped by 1.21 and the mean F1-score decreased by 1.44 after this change. This indicates that low-quality samples may reduce the performance of the whistle extraction network training, and our simple heuristic selection method effectively selects samples for the whistle extraction task. 

\begin{figure}[t!]
  \centering
  \includegraphics[width=0.9\linewidth]{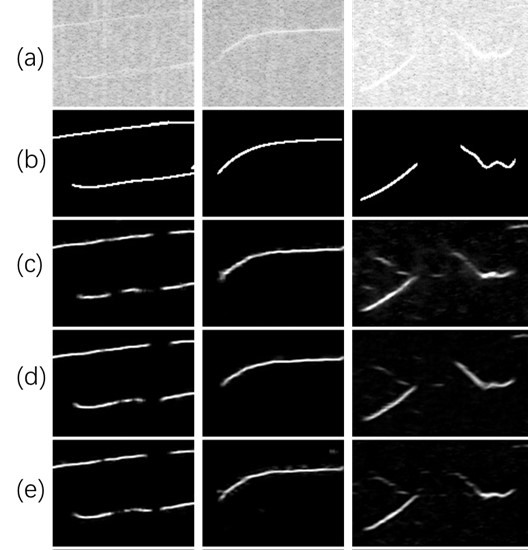}
  \caption{Outputs of whistle extraction models. Models with the best whistle extraction F1-score among all parallel experiments in each training setting are visualized. (a) Spectrograms that are used as model input. (b) Ground truth. (c) Output of model trained with 2500 real positive patches and negative patches. (d) Output of model trained with 2500 positive patches and negative patches and GAN synthesized data. (e) Same as d, but the model does not have auxiliary batch normalization (ABN).}
  \label{fig:fig_whistle_ablation}
\end{figure}

\subsubsection{ABN}
Because ABN stores statistics of real samples and generated samples separately, it may better stabilize the training when the generated samples and real samples have different distributions \cite{xie2020adversarial}. We evaluated the functionality of applying ABN with and without patch selection to our whistle extraction task; patch selection affects the generated sample distribution. After removal of ABN, the whistle extraction F1-score dropped by 0.98 with patch selection and by 1.97 without patch selection. This suggests that our patch selection method contributes to generating samples that are closer to the actual distribution of whistles. The performance change is consistent with our hypothesis that generated samples and real samples have a different distribution when few data are included in GAN training. 

We also observed decreases in ODS of 0.68 with patch selection and 0.86 without patch selection after removal of ABN, which is a less decrease compared to whistle extraction F1-score. 
% Pu: Remove those description for updated results. 
% which is a less significant decrease compared to whistle extraction F1-score. 
% Although the ODS measure is consistent with the whistle extraction F1-score when we added more real samples (Section~\ref{sec:data_aug_results}), adding generated samples may have different effects. 
While ODS demonstrates the whistle extraction model’s performance at the spectrogram bin level, this metric does not always linearly correlate to the whistle extraction performance, because it ignores the signal continuity among bins. 
% the contour extraction system is designed to handle occasional missing peaks. 
We observed that removing ABN frequently resulted in poorer continuity of predicted patches (e.g., Fig.~\ref{fig:fig_whistle_ablation}d and \ref{fig:fig_whistle_ablation}e, first and third examples) and a greater number of false positives (e.g., Fig.~\ref{fig:fig_whistle_ablation}d and \ref{fig:fig_whistle_ablation}e, second example). The whistle extraction F1-score also indicates the model’s ability to recognize whistle signals under varying noise conditions or suppress false positives in the high-energy region of spectrogram according to the context information (signals in the neighborhood). The generated whistle contour and signals may be less continuous than the real samples, which will train the whistle extraction model to ignore context information and make discontinuous predictions when ABN is removed. 
% Pu: remove the following discussion. 
% While those discontinuous predictions may affect whistle extraction F1-score more than ODS, this may explain why we have less significant decrease in ODS than the decrease in F1-score in this ablation study. 
% This may also indicate that use of ABN is necessary to improve whistle extraction results. 
The comparison among  Fig.~\ref{fig:fig_whistle_ablation} (c), (d), (e) rightmost column also shows that use of our generated data reduces false positives. 

\subsubsection{Stage-wise GAN}
Instead of decomposing the sample generation into multiple stages, we used a single WGAN-gp with two output channels to generate whistle data, the spectrogram samples and their labels, similar to \cite{bowles2018gan}. To deal with the increased learning difficulty of one WGAN, we increased the WGAN-gp capacity of the generator by using twice the number of hidden layers for each convolutional layer output as that in Section~\ref{sec:wgan}. Examples of samples generated by this model are shown in Fig.~\ref{fig:fig_whistle_vanilla}. We saw clear artifacts and unnatural, sudden changes in the magnitude in adjacent bins on the spectrogram. The visual quality of generated samples was substantially worse than those generated by our stage-wise GAN in Fig.~\ref{fig:whistle_gan_examples_detail}. We also observed a decrease of 1.04 in the whistle extraction F1-score compared to our proposed framework. Data augmentation with the low-quality samples still permitted the performance of the model to surpass that without augmentation for the time-frequency detection task. The negative effect of using corrupted data might be mitigated by the ABN layer.

\subsubsection{The third GAN}
In this ablation study, we remove the third GAN and instead generate positive sample $I_P'$ by simply adding the generated whistle contour $I_W$ to the generated background noise $I_N$. Following the work of Li et al. \cite{li2020learning}, we apply Gaussian blur $G$ with random deviation parameter $\sigma$ to the whistle contour, and we add the blurred contour to the background noise: 
\begin{equation}
\label{exp:ab_addition_blur}
I_P'=I_N+\lambda CLIP(I_W+G(Y,\ \sigma))
% X\prime=X+\lambda CLIP(Y+G(Y,\ \sigma))
\end{equation}
where the clipping function $CLIP(x)$ is 
\begin{equation}
\label{exp:ab_addition_clip}
CLIP(x)=\begin{cases}
 &0 \text{, } x\in(-\infty, 0) \\
 &x \text{, } x\in[0,1] \\
 &1 \text{,  } x\in(1, +\infty)
\end{cases}
\end{equation}
We also try a simple version which does not contain Gaussian blur:
\begin{equation}
\label{exp:ab_addition}
% X\prime=X+\lambda Y
I_P'=I_N+\lambda I_W
\end{equation}
where $\lambda$ is a random weighting parameter.
We use the same parameter setting as Li et al. \cite{li2020learning}, where $\lambda$ and $\sigma$ are uniform random numbers within the ranges of $[0.03,\ 0.23]$ and  $[0.3,\ 1.3]$, respectively. As shown in Table~\ref{tab:gan_ablation}, both methods in Equation~\ref{exp:ab_addition_clip} and Equation~\ref{exp:ab_addition} lead to inferior performance compared to the proposed stage-wise GAN method (``2500+GAN") that uses the third GAN. Considering that we use the same set of background noise and whistle contour shapes, this ablation study indicates that our proposed stage-wise GAN method generates more realistic whistle signals with a  appearance which contributes to the improved training of the whistle model.

\begin{table}[t!]
\caption{Comparison of whistle extraction methods}
\label{tab:gan_ablation}
\begin{tabular}{c|c|c|c}
\hline
Method                    & F1-score & Precision & Recall \\ \hline
Roch   et al., 2011 \cite{roch2011automated}      & 75.95    & 81.125    & 72.275 \\ \hline
Gruden et al., 2020 \cite{gruden2020automated}      & 83.40    & 76.55     & 92.45  \\ \hline
Gruden et al., 2020  ($\geq$150ms) & 74.38 & 95.85  & 60.875 \\ \hline
Li et al., 2020  \cite{li2020learning}           & 87.42    & 89.27     & 86.04  \\ \hline
Li et al., 2020 + our GAN & 87.88    & 89.60     & 86.63  \\ \hline
\end{tabular}
\end{table}

\subsection{Comparison with other whistle extraction methods}
%In addition to our previous work on network-based whistle extraction \cite{li2020learning}, 
%we select two representative and competitive whistle extraction methods for comparison. Both methods generate whistle candidate points by checking if SNR values are above a threshold on the denoised spectrogram. The Graph-Search method from Roch et al. \cite{roch2011automated} maintains graphs consisting of candidate points, which are extended with new points on the basis of how well those new candidate points concur with the existing graph based on polynomial fitting results. As a comparison, Gruden et al.\cite{gruden2020automated} uses a probabilistic approach based on sequential Monte-Carlo probability hypothesis density (SMCPHD). In addition to the result of all SMCPHD predictions,  we also present the result of predictions that are longer than 150ms, because Graph-Search and our method apply these length criteria for detection.

In addition to our previous work on network-based whistle extraction \cite{li2020learning}, we have selected two representative and competitive whistle extraction methods for comparison. Both methods identify whistle candidate points by determining if the Signal-to-Noise Ratio (SNR) values are above a threshold on the denoised spectrogram. The Graph-Search method developed by Roch et al. \cite{roch2011automated} employs graphs consisting of candidate points, which are extended with new points based on how well these new candidate points align with the existing graph through polynomial fitting results. As a point of comparison, Gruden et al. \cite{gruden2020automated} uses a probabilistic approach based on the sequential Monte-Carlo probability hypothesis density (SMCPHD). In addition to the result of all SMCPHD predictions, we also present the results of predictions that are longer than 150ms, as both Graph-Search and our method apply this length criterion for detection.

Our approach outperforms SMCPHD and Graph-Search in the whistle extraction F1-score by 4.48 and 11.93, respectively. Additionally, our GAN-generated samples improve the method in \cite{li2020learning} by 0.46 in F1-score, 0.33 in precision, and 0.59 in recall. SMCPHD demonstrates the highest recall but the lowest precision in this comparison, which indicates its aggressive strategy of making more whistle predictions. By removing whistle detections by SMCPHD that are shorter than 150ms, the precision of SMCPHD is improved by 19.3, while the recall is decreased by 31.57. This study suggests that SMCPHD prefers shorter segments of whistles in its predictions. Our GAN-generated samples help the learning-based model achieve a competitive performance advantage on this whistle extraction task, however, it should be noted that optimizing the other algorithms for this specific dataset may diminish these advantages.

%Our approach surpasses the whistle extraction F1-score of SMCPHD and Graph-Search by 4.48 and 11.93, respectively, while our GAN-generated samples improve the method in \cite{li2020learning} by 0.46 in F1-score, 0.33 in precision, and 0.59 in recall. SMCPHD produces the highest recall but lowest precision in this comparison, which indicates an aggressive strategy that results in more whistle predictions. When we apply 150ms length criteria to SMCPHD predictions, the precision of SMCPHD is improved by 19.3 while the recall is decreased by 31.57, which indicates that SMCPHD prefer shorter segments of whistles in its predictions. Our GAN-generated samples help the learning-based model achieve a competitive performance advantage on this whistle extraction task, but one should remember that optimizing the other algorithms for this specific dataset could diminish these advantages.

\section{CONCLUSION AND DISCUSSION}
We present a framework of stage-wise generative adversarial networks to generate training samples for whistle extraction. The data generation process consists of three stages: (i) generate time x frequency spectrogram patches containing background noise (ii) generate whistle contours and automatically discard poor quality contours (iii) fuse whistle signals with the background noise. Each stage is completed by one trained generative adversarial network. Compared to using a single vanilla GAN generating whistle extraction data and labels, our stage-wise GANs can generate samples with fewer artifacts which results in increased whistle extraction performance. We examined our data generation method by a series of experiments employing differing quantities of real and generated data, and note that using the generated data lead to consistent performance gains. 

The stage-wise design may mainly contribute to the success of our data generation method. It separates the modeling of different components and the relationship between components, which eases the learning of the GANs in each stage as well as provides a straightforward way to explore different combinations of components. In our case, we generated the background noise separately and we were able to add different whistle signals to the same background. If we directly apply this idea to semantic segmentation data generation of natural images, we may first generate the appearance of background scene, then generate objects on it according to a desired segmentation map. If we extend this idea, we may generate the appearance of different objects separately and then add them to the background. In this way, we may fully explore combinations of varying objects and background appearances in the same segmentation layout. In our whistle extraction experiments, we did not use this extended idea, because the appearance of our foreground object (the whistles) is relatively simple, i.e., the variance of appearance is mainly rooted in the whistle contour shape and whistle magnitude. Therefore, we directly add whistle signals to the background using the third GAN in our framework. Our framework can be readily extended to extract calls of other whale species (e.g, baleen whales) and to other similar tasks (e.g., semantic image segmentation).

Though it may not affect the main contributions of this work, our data generation method can be improved in three aspects in the future. Firstly, we may use improved generative neural network architecture and training strategies. For example, we may use a generator architecture based on a style-transfer network which improves the generated sample quality \cite{karras2019style}. The discriminator augmentation mechanism proposed in \cite{karras2020training} may help stabilize training in limited data regimes. We may also explore generating larger patches of high quality with the method in \cite{karras2018progressive}. Secondly, we may use real data in the data generation process to enrich the data variance. The real background and annotated whistle contours can be used as the input data of our GAN in the third stage, and we can generate whistle signsals of novel shapes on real background or generate whistle signals of annotated contour shapes on GAN-generated background. Thirdly, we may improve the sample selection method. In this paper, we use a simple yet effective pixel-wise entropy method to select whistle contour of good quality. Metric measuring texture or semantic information like \cite{larkin2016reflections} may better measure the quality of our generated samples and improve the sample selection process.

\section*{Acknowledgments}
Thanks to John A. Hildebrand and Simone Baumann-Pickering of Scripps Institution of Oceanography and Melissa S. Soldevilla of the National Oceanic and Atmospheric Administration (NOAA) for providing the acoustic data to the DCLDE 2011 organizing committee. We appreciate the effort of Shannon Rankin and Yvonne Barkley of NOAA in producing portions of the DCLDE 2011 annotations. We thank Michael Weise of the US Office of Naval Research for the support (N000141712867 and N000142112567).

\bibliographystyle{IEEEtran}
\bibliography{ref}

% Generated by IEEEtran.bst, version: 1.14 (2015/08/26)
\begin{thebibliography}{10}
\providecommand{\url}[1]{#1}
\csname url@samestyle\endcsname
\providecommand{\newblock}{\relax}
\providecommand{\bibinfo}[2]{#2}
\providecommand{\BIBentrySTDinterwordspacing}{\spaceskip=0pt\relax}
\providecommand{\BIBentryALTinterwordstretchfactor}{4}
\providecommand{\BIBentryALTinterwordspacing}{\spaceskip=\fontdimen2\font plus
\BIBentryALTinterwordstretchfactor\fontdimen3\font minus
  \fontdimen4\font\relax}
\providecommand{\BIBforeignlanguage}[2]{{%
\expandafter\ifx\csname l@#1\endcsname\relax
\typeout{** WARNING: IEEEtran.bst: No hyphenation pattern has been}%
\typeout{** loaded for the language `#1'. Using the pattern for}%
\typeout{** the default language instead.}%
\else
\language=\csname l@#1\endcsname
\fi
#2}}
\providecommand{\BIBdecl}{\relax}
\BIBdecl

\bibitem{rabiner1993fundamentals}
L.~Rabiner and B.-H. Juang, \emph{Fundamentals of Speech Recognition}.\hskip
  1em plus 0.5em minus 0.4em\relax Prentice-Hall, Inc., 1993.

\bibitem{ren2016sound}
J.~Ren, X.~Jiang, J.~Yuan, and N.~Magnenat-Thalmann, ``Sound-event
  classification using robust texture features for robot hearing,'' \emph{IEEE
  Transactions on Multimedia}, vol.~19, no.~3, pp. 447--458, 2016.

\bibitem{kahl2021birdnet}
S.~Kahl, C.~M. Wood, M.~Eibl, and H.~Klinck, ``{BirdNET}: A deep learning
  solution for avian diversity monitoring,'' \emph{Ecological Informatics},
  vol.~61, p. 101236, 2021.

\bibitem{lee2009automatic}
C.-H. Lee, J.-L. Shih, K.-M. Yu, and H.-S. Lin, ``Automatic music genre
  classification based on modulation spectral analysis of spectral and cepstral
  features,'' \emph{IEEE Transactions on Multimedia}, vol.~11, no.~4, pp.
  670--682, 2009.

\bibitem{rizzi2017instrument}
A.~Rizzi, M.~Antonelli, and M.~Luzi, ``Instrument learning and sparse nmd for
  automatic polyphonic music transcription,'' \emph{IEEE Transactions on
  Multimedia}, vol.~19, no.~7, pp. 1405--1415, 2017.

\bibitem{zhang2017speech}
S.~Zhang, S.~Zhang, T.~Huang, and W.~Gao, ``Speech emotion recognition using
  deep convolutional neural network and discriminant temporal pyramid
  matching,'' \emph{IEEE Transactions on Multimedia}, vol.~20, no.~6, pp.
  1576--1590, 2017.

\bibitem{yost2001fundamentals}
W.~A. Yost, ``Fundamentals of hearing: An introduction,'' 2001.

\bibitem{vijayan2018analysis}
K.~Vijayan, X.~Gao, and H.~Li, ``Analysis of speech and singing signals for
  temporal alignment,'' in \emph{2018 Asia-Pacific Signal and Information
  Processing Association Annual Summit and Conference (APSIPA ASC)}.\hskip 1em
  plus 0.5em minus 0.4em\relax IEEE, 2018, pp. 1893--1898.

\bibitem{met2021spectrogram}
B.~Met-Montot, S.~Cabon, G.~Carrault, and F.~Por{\'e}e, ``Spectrogram-based
  fundamental frequency tracking of spontaneous cries in preterm newborns,'' in
  \emph{2020 28th European Signal Processing Conference (EUSIPCO)}.\hskip 1em
  plus 0.5em minus 0.4em\relax IEEE, 2021, pp. 1185--1189.

\bibitem{lu2018vocal}
W.~T. Lu, L.~Su \emph{et~al.}, ``Vocal melody extraction with semantic
  segmentation and audio-symbolic domain transfer learning.'' in
  \emph{International Society for Music Information Retrieval (ISMIR)}, 2018,
  pp. 521--528.

\bibitem{roch2011automated}
M.~A. Roch, T.~Scott~Brandes, B.~Patel, Y.~Barkley, S.~Baumann-Pickering, and
  M.~S. Soldevilla, ``Automated extraction of odontocete whistle contours,''
  \emph{The Journal of the Acoustical Society of America}, vol. 130, no.~4, pp.
  2212--2223, 2011.

\bibitem{gillespie2013automatic}
D.~Gillespie, M.~Caillat, J.~Gordon, and P.~White, ``Automatic detection and
  classification of odontocete whistles,'' \emph{The Journal of the Acoustical
  Society of America}, vol. 134, no.~3, pp. 2427--2437, 2013.

\bibitem{jiang2019whistle}
J.-j. Jiang, L.-r. Bu, F.-j. Duan, X.-q. Wang, W.~Liu, Z.-b. Sun, and C.-y. Li,
  ``Whistle detection and classification for whales based on convolutional
  neural networks,'' \emph{Applied Acoustics}, vol. 150, pp. 169--178, 2019.

\bibitem{janik2013identifying}
V.~M. Janik, S.~L. King, L.~S. Sayigh, and R.~S. Wells, ``Identifying signature
  whistles from recordings of groups of unrestrained bottlenose dolphins
  (\textit{Tursiops truncatus}),'' \emph{Marine Mammal Science}, vol.~29,
  no.~1, pp. 109--122, 2013.

\bibitem{taruski1979whistle}
A.~G. Taruski, ``The whistle repertoire of the north atlantic pilot whale
  (\textit{Globicephala melaena}) and its relationship to behavior and
  environment,'' in \emph{Behavior of marine animals}.\hskip 1em plus 0.5em
  minus 0.4em\relax Springer, 1979, pp. 345--368.

\bibitem{sjare1986relationship}
B.~L. Sjare and T.~G. Smith, ``The relationship between behavioral activity and
  underwater vocalizations of the white whale, \textit{Delphinapterus
  leucas},'' \emph{Canadian Journal of Zoology}, vol.~64, no.~12, pp.
  2824--2831, 1986.

\bibitem{thomsen2002communicative}
F.~Thomsen, D.~Franck, and J.~K. Ford, ``On the communicative significance of
  whistles in wild killer whales (\textit{Orcinus orca}),''
  \emph{Naturwissenschaften}, vol.~89, no.~9, pp. 404--407, 2002.

\bibitem{jaramillo2017passive}
A.~Jaramillo-Legorreta, G.~Cardenas-Hinojosa, E.~Nieto-Garcia, L.~Rojas-Bracho,
  J.~Ver~Hoef, J.~Moore, N.~Tregenza, J.~Barlow, T.~Gerrodette, L.~Thomas
  \emph{et~al.}, ``Passive acoustic monitoring of the decline of mexico's
  critically endangered vaquita,'' \emph{Conservation Biology}, vol.~31, no.~1,
  pp. 183--191, 2017.

\bibitem{li2020learning}
P.~Li, X.~Liu, K.~Palmer, E.~Fleishman, D.~Gillespie, E.-M. Nosal, Y.~Shiu,
  H.~Klinck, D.~Cholewiak, T.~Helble \emph{et~al.}, ``Learning deep models from
  synthetic data for extracting dolphin whistle contours,'' in \emph{2020
  International Joint Conference on Neural Networks (IJCNN)}.\hskip 1em plus
  0.5em minus 0.4em\relax IEEE, 2020, pp. 1--10.

\bibitem{mallawaarachchi2008spectrogram}
A.~Mallawaarachchi, S.~Ong, M.~Chitre, and E.~Taylor, ``Spectrogram denoising
  and automated extraction of the fundamental frequency variation of dolphin
  whistles,'' \emph{The Journal of the Acoustical Society of America}, vol.
  124, no.~2, pp. 1159--1170, 2008.

\bibitem{chen2018sparse}
H.~Chen, J.~Yan, N.~U.~R. Junejo, J.~Qi, and H.~Sun, ``Sparse representation
  based on tunable q-factor wavelet transform for whale click and whistle
  extraction,'' \emph{Shock and Vibration}, vol. 2018, 2018.

\bibitem{gruden2020automated}
P.~Gruden and P.~R. White, ``Automated extraction of dolphin whistles—a
  sequential {Monte Carlo} probability hypothesis density approach,'' \emph{The
  Journal of the Acoustical Society of America}, vol. 148, no.~5, pp.
  3014--3026, 2020.

\bibitem{wang2021method}
X.~Wang, J.~Jiang, F.~Duan, C.~Liang, C.~Li, Z.~Sun, R.~Lu, F.~Li, J.~Xu, and
  X.~Fu, ``A method for enhancement and automated extraction and tracing of
  \textit{odontoceti} whistle signals base on time-frequency spectrogram,''
  \emph{Applied Acoustics}, vol. 176, p. 107698, 2021.

\bibitem{jeong2019consistency}
J.~Jeong, S.~Lee, J.~Kim, and N.~Kwak, ``Consistency-based semi-supervised
  learning for object detection,'' \emph{Advances in Neural Information
  Processing Systems}, vol.~32, 2019.

\bibitem{li2021learning}
P.~Li and X.~Liu, ``Learning knowledge-rich sequential model for planar
  homography estimation in aerial video,'' in \emph{2020 25th International
  Conference on Pattern Recognition (ICPR)}.\hskip 1em plus 0.5em minus
  0.4em\relax IEEE, 2021, pp. 10\,584--10\,591.

\bibitem{he2016deep}
K.~He, X.~Zhang, S.~Ren, and J.~Sun, ``Deep residual learning for image
  recognition,'' in \emph{Proceedings of the IEEE conference on Computer Vision
  and Pattern Recognition}, 2016, pp. 770--778.

\bibitem{dao2019kernel}
T.~Dao, A.~Gu, A.~Ratner, V.~Smith, C.~De~Sa, and C.~R{\'e}, ``A kernel theory
  of modern data augmentation,'' in \emph{International Conference on Machine
  Learning}.\hskip 1em plus 0.5em minus 0.4em\relax PMLR, 2019, pp. 1528--1537.

\bibitem{brumm2011evolution}
H.~Brumm and S.~A. Zollinger, ``The evolution of the {Lombard} effect: 100
  years of psychoacoustic research,'' \emph{Behaviour}, vol. 148, no. 11-13,
  pp. 1173--1198, 2011.

\bibitem{au1985demonstration}
W.~W. Au, D.~A. Carder, R.~H. Penner, and B.~L. Scronce, ``Demonstration of
  adaptation in beluga whale echolocation signals,'' \emph{The Journal of the
  Acoustical Society of America}, vol.~77, no.~2, pp. 726--730, 1985.

\bibitem{zhu2017unpaired}
J.-Y. Zhu, T.~Park, P.~Isola, and A.~A. Efros, ``Unpaired image-to-image
  translation using cycle-consistent adversarial networks,'' in
  \emph{Proceedings of the IEEE international conference on computer vision},
  2017, pp. 2223--2232.

\bibitem{xie2020adversarial}
C.~Xie, M.~Tan, B.~Gong, J.~Wang, A.~L. Yuille, and Q.~V. Le, ``Adversarial
  examples improve image recognition,'' in \emph{Proceedings of the IEEE/CVF
  Conference on Computer Vision and Pattern Recognition}, 2020, pp. 819--828.

\bibitem{wali2022generative}
A.~Wali, Z.~Alamgir, S.~Karim, A.~Fawaz, M.~B. Ali, M.~Adan, and M.~Mujtaba,
  ``Generative adversarial networks for speech processing: A review,''
  \emph{Computer Speech \& Language}, vol.~72, p. 101308, 2022.

\bibitem{dadouchi2013automated}
F.~Dadouchi, C.~Gervaise, C.~Ioana, J.~Huillery, and J.~I. Mars, ``Automated
  segmentation of linear time-frequency representations of marine-mammal
  sounds,'' \emph{The Journal of the Acoustical Society of America}, vol. 134,
  no.~3, pp. 2546--2555, 2013.

\bibitem{halkias2006call}
X.~C. Halkias and D.~P. Ellis, ``Call detection and extraction using bayesian
  inference,'' \emph{Applied Acoustics}, vol.~67, no. 11-12, pp. 1164--1174,
  2006.

\bibitem{white2008introduction}
P.~White and M.~Hadley, ``Introduction to particle filters for tracking
  applications in the passive acoustic monitoring of cetaceans,''
  \emph{Canadian Acoustics}, vol.~36, no.~1, pp. 146--152, 2008.

\bibitem{gruden2016automated}
P.~Gruden and P.~R. White, ``Automated tracking of dolphin whistles using
  gaussian mixture probability hypothesis density filters,'' \emph{The Journal
  of the Acoustical Society of America}, vol. 140, no.~3, pp. 1981--1991, 2016.

\bibitem{mellinger2011method}
D.~K. Mellinger, S.~W. Martin, R.~P. Morrissey, L.~Thomas, and J.~J. Yosco, ``A
  method for detecting whistles, moans, and other frequency contour sounds,''
  \emph{The Journal of the Acoustical Society of America}, vol. 129, no.~6, pp.
  4055--4061, 2011.

\bibitem{kershenbaum2013image}
A.~Kershenbaum and M.~A. Roch, ``An image processing based paradigm for the
  extraction of tonal sounds in cetacean communications,'' \emph{The Journal of
  the Acoustical Society of America}, vol. 134, no.~6, pp. 4435--4445, 2013.

\bibitem{serra2020active}
O.~Serra, F.~Martins, and L.~R. Padovese, ``Active contour-based detection of
  estuarine dolphin whistles in spectrogram images,'' \emph{Ecological
  Informatics}, vol.~55, p. 101036, 2020.

\bibitem{han2014neural}
K.~Han and D.~Wang, ``Neural network based pitch tracking in very noisy
  speech,'' \emph{IEEE/ACM Transactions on Audio, Speech, and Language
  Processing}, vol.~22, no.~12, pp. 2158--2168, 2014.

\bibitem{bittner2017deep}
R.~M. Bittner, B.~McFee, J.~Salamon, P.~Li, and J.~P. Bello, ``Deep salience
  representations for {$F_0$} estimation in polyphonic music.'' in
  \emph{International Society for Music Information Retrieval (ISMIR)}, 2017,
  pp. 63--70.

\bibitem{liu2018classification}
S.~Liu, M.~Liu, M.~Wang, T.~Ma, and X.~Qing, ``Classification of cetacean
  whistles based on convolutional neural network,'' in \emph{2018 10th
  International Conference on Wireless Communications and Signal Processing
  (WCSP)}.\hskip 1em plus 0.5em minus 0.4em\relax IEEE, 2018, pp. 1--5.

\bibitem{ngo2021self}
M.~Ngo, S.~Karaoglu, and T.~Gevers, ``Self-supervised face image manipulation
  by conditioning {GAN} on face decomposition,'' \emph{IEEE Transactions on
  Multimedia}, 2021.

\bibitem{galteri2019deep}
L.~Galteri, L.~Seidenari, M.~Bertini, and A.~Del~Bimbo, ``Deep universal
  generative adversarial compression artifact removal,'' \emph{IEEE
  Transactions on Multimedia}, vol.~21, no.~8, pp. 2131--2145, 2019.

\bibitem{lu2021generate}
J.~Lu, W.~Zhang, and H.~Yin, ``Generate and purify: Efficient person data
  generation for re-identification,'' \emph{IEEE Transactions on Multimedia},
  2021.

\bibitem{goodfellow2014generative}
I.~Goodfellow, J.~Pouget-Abadie, M.~Mirza, B.~Xu, D.~Warde-Farley, S.~Ozair,
  A.~Courville, and Y.~Bengio, ``Generative adversarial nets,'' \emph{Advances
  in Neural Information Processing Systems}, vol.~27, 2014.

\bibitem{radford2015unsupervised}
A.~Radford, L.~Metz, and S.~Chintala, ``Unsupervised representation learning
  with deep convolutional generative adversarial networks,'' \emph{arXiv
  preprint arXiv:1511.06434}, 2015.

\bibitem{gulrajani2017improved}
I.~Gulrajani, F.~Ahmed, M.~Arjovsky, V.~Dumoulin, and A.~C. Courville,
  ``Improved training of {Wasserstein} {GANs},'' \emph{Advances in Neural
  Information Processing Systems}, vol.~30, 2017.

\bibitem{emami2020spa}
H.~Emami, M.~M. Aliabadi, M.~Dong, and R.~B. Chinnam, ``{SPA-GAN}: Spatial
  attention {GAN} for image-to-image translation,'' \emph{IEEE Transactions on
  Multimedia}, vol.~23, pp. 391--401, 2020.

\bibitem{chen2019quality}
L.~Chen, L.~Wu, Z.~Hu, and M.~Wang, ``Quality-aware unpaired image-to-image
  translation,'' \emph{IEEE Transactions on Multimedia}, vol.~21, no.~10, pp.
  2664--2674, 2019.

\bibitem{frid2018synthetic}
M.~Frid-Adar, E.~Klang, M.~Amitai, J.~Goldberger, and H.~Greenspan, ``Synthetic
  data augmentation using {GAN} for improved liver lesion classification,'' in
  \emph{2018 IEEE 15th international symposium on biomedical imaging (ISBI
  2018)}.\hskip 1em plus 0.5em minus 0.4em\relax IEEE, 2018, pp. 289--293.

\bibitem{mariani2018bagan}
G.~Mariani, F.~Scheidegger, R.~Istrate, C.~Bekas, and C.~Malossi, ``{BAGAN}:
  Data augmentation with balancing {GAN},'' \emph{arXiv preprint
  arXiv:1803.09655}, 2018.

\bibitem{bowles2018gan}
C.~Bowles, L.~Chen, R.~Guerrero, P.~Bentley, R.~Gunn, A.~Hammers, D.~A. Dickie,
  M.~V. Hern{\'a}ndez, J.~Wardlaw, and D.~Rueckert, ``{GAN} augmentation:
  Augmenting training data using generative adversarial networks,'' \emph{arXiv
  preprint arXiv:1810.10863}, 2018.

\bibitem{huang2018auggan}
S.-W. Huang, C.-T. Lin, S.-P. Chen, Y.-Y. Wu, P.-H. Hsu, and S.-H. Lai,
  ``{AugGAN}: Cross domain adaptation with {GAN}-based data augmentation,'' in
  \emph{Proceedings of the European Conference on Computer Vision (ECCV)},
  2018, pp. 718--731.

\bibitem{zeng2021strokegan}
J.~Zeng, Q.~Chen, Y.~Liu, M.~Wang, and Y.~Yao, ``{StrokeGAN}: Reducing mode
  collapse in chinese font generation via stroke encoding,'' in
  \emph{Proceedings of the AAAI Conference on Artificial Intelligence},
  vol.~35, no.~4, 2021, pp. 3270--3277.

\bibitem{osakabe2021cyclegan}
T.~Osakabe, M.~Tanaka, Y.~Kinoshita, and H.~Kiya, ``{CycleGAN} without
  checkerboard artifacts for counter-forensics of fake-image detection,'' in
  \emph{International Workshop on Advanced Imaging Technology (IWAIT) 2021},
  vol. 11766.\hskip 1em plus 0.5em minus 0.4em\relax International Society for
  Optics and Photonics, 2021, p. 1176609.

\bibitem{mun2017generative}
S.~Mun, S.~Park, D.~K. Han, and H.~Ko, ``Generative adversarial network based
  acoustic scene training set augmentation and selection using {SVM}
  hyper-plane,'' \emph{Proceedings of the Detection and Classification of
  Acoustic Scenes and Events (DCASE)}, pp. 93--97, 2017.

\bibitem{agarwal2021detecting}
S.~Agarwal and H.~Farid, ``Detecting deep-fake videos from aural and oral
  dynamics,'' in \emph{Proceedings of the IEEE/CVF Conference on Computer
  Vision and Pattern Recognition}, 2021, pp. 981--989.

\bibitem{antoniou2017data}
A.~Antoniou, A.~Storkey, and H.~Edwards, ``Data augmentation generative
  adversarial networks,'' \emph{arXiv preprint arXiv:1711.04340}, 2017.

\bibitem{mu2020learning}
J.~Mu, W.~Qiu, G.~D. Hager, and A.~L. Yuille, ``Learning from synthetic
  animals,'' in \emph{Proceedings of the IEEE/CVF Conference on Computer Vision
  and Pattern Recognition}, 2020, pp. 12\,386--12\,395.

\bibitem{waheed2020covidgan}
A.~Waheed, M.~Goyal, D.~Gupta, A.~Khanna, F.~Al-Turjman, and P.~R. Pinheiro,
  ``{CovidGAN}: data augmentation using auxiliary classifier {GAN} for improved
  {COVID-19} detection,'' \emph{IEEE Access}, vol.~8, pp. 91\,916--91\,923,
  2020.

\bibitem{pandey2020image}
S.~Pandey, P.~R. Singh, and J.~Tian, ``An image augmentation approach using
  two-stage generative adversarial network for nuclei image segmentation,''
  \emph{Biomedical Signal Processing and Control}, vol.~57, p. 101782, 2020.

\bibitem{mellinger2006mobysound}
\BIBentryALTinterwordspacing
D.~K. Mellinger and C.~W. Clark, ``Mobysound: A reference archive for studying
  automatic recognition of marine mammal sounds,'' \emph{Applied Acoustics},
  vol.~67, no. 11-12, pp. 1226--1242, 2006, (Webpage last viewed on May 23,
  2022). [Online]. Available: \url{http://www.mobysound.org/workshops_p2.html}
\BIBentrySTDinterwordspacing

\bibitem{ronneberger2015u}
O.~Ronneberger, P.~Fischer, and T.~Brox, ``{U-Net}: Convolutional networks for
  biomedical image segmentation,'' in \emph{International Conference on Medical
  image computing and computer-assisted intervention}.\hskip 1em plus 0.5em
  minus 0.4em\relax Springer, 2015, pp. 234--241.

\bibitem{arbelaez2010contour}
P.~Arbelaez, M.~Maire, C.~Fowlkes, and J.~Malik, ``Contour detection and
  hierarchical image segmentation,'' \emph{IEEE transactions on pattern
  analysis and machine intelligence}, vol.~33, no.~5, pp. 898--916, 2010.

\bibitem{althnian2021impact}
A.~Althnian, D.~AlSaeed, H.~Al-Baity, A.~Samha, A.~B. Dris, N.~Alzakari,
  A.~Abou~Elwafa, and H.~Kurdi, ``Impact of dataset size on classification
  performance: an empirical evaluation in the medical domain,'' \emph{Applied
  Sciences}, vol.~11, no.~2, p. 796, 2021.

\bibitem{karras2019style}
T.~Karras, S.~Laine, and T.~Aila, ``A style-based generator architecture for
  generative adversarial networks,'' in \emph{Proceedings of the IEEE/CVF
  conference on Computer Vision and Pattern Recognition}, 2019, pp. 4401--4410.

\bibitem{karras2020training}
T.~Karras, M.~Aittala, J.~Hellsten, S.~Laine, J.~Lehtinen, and T.~Aila,
  ``Training generative adversarial networks with limited data,''
  \emph{Advances in Neural Information Processing Systems}, vol.~33, pp.
  12\,104--12\,114, 2020.

\bibitem{karras2018progressive}
T.~Karras, T.~Aila, S.~Laine, and J.~Lehtinen, ``Progressive growing of {GANs}
  for improved quality, stability, and variation,'' in \emph{International
  Conference on Learning Representations}, 2018.

\bibitem{larkin2016reflections}
K.~G. Larkin, ``Reflections on {Shannon Information}: In search of a natural
  information-entropy for images,'' \emph{arXiv preprint arXiv:1609.01117},
  2016.

\end{thebibliography}

\vfill
\end{document}